%% file: main.tex
\documentclass[10pt,twocolumn,letterpaper]{article}

\usepackage[pagenumbers]{cvpr} %

\input{preamble}

\input{macros}

\definecolor{cvprblue}{rgb}{0.21,0.49,0.74}

\usepackage[pagebackref,breaklinks,colorlinks,allcolors=cvprblue]{hyperref}
\usepackage{multirow}
\usepackage{bm}
\usepackage{amsmath}
\usepackage{amssymb}
\usepackage{adjustbox}
\usepackage{makecell}
\usepackage{xspace}
\usepackage{pifont}
\usepackage{bbding} 
\usepackage{animate}
\usepackage{varwidth}
\usepackage{microtype} 
\definecolor{darkgreen}{rgb}{0,0.5,0}
\usepackage{colortbl}
\definecolor{lightgray}{rgb}{0.8, 0.8, 0.8} %
\newcommand{\cmark}{\textcolor{darkgreen}{\ding{51}}}
\newcommand{\xmark}{\textcolor{red}{\ding{55}}}%
\usepackage{bm}

\def\paperID{4877} %
\def\confName{CVPR}
\def\confYear{2025}

\title{\includegraphics[scale=0.15]{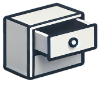}DRAWER: Digital Reconstruction and Articulation With Environment Realism}

\author{
Hongchi Xia$^{1}$
\enspace
Entong Su$^{2}$
\enspace
Marius Memmel$^{2}$
\enspace
Arhan Jain$^{2}$
\enspace
Raymond Yu$^{2}$
\enspace
Numfor Mbiziwo-Tiapo$^{2}$\\
Ali Farhadi$^{2,3}$
\quad 
Abhishek Gupta$^{2}$
\quad 
Shenlong Wang$^{1}$
\quad 
Wei-Chiu Ma$^{4}$
\vspace{2mm}\\
$^{1}$University of Illinois Urbana-Champaign
\quad
$^{2}$University of Washington
\\
$^{3}$Allen Institute for AI \quad 
$^{4}$Cornell University\\
}

\def\modelname{DRAWER}

\begin{document}

\twocolumn[{%
\renewcommand\twocolumn[1][]{#1}%
\maketitle
\vspace{-13mm}
\captionsetup{type=figure}
\begin{center}
\includegraphics[width=0.95\linewidth]{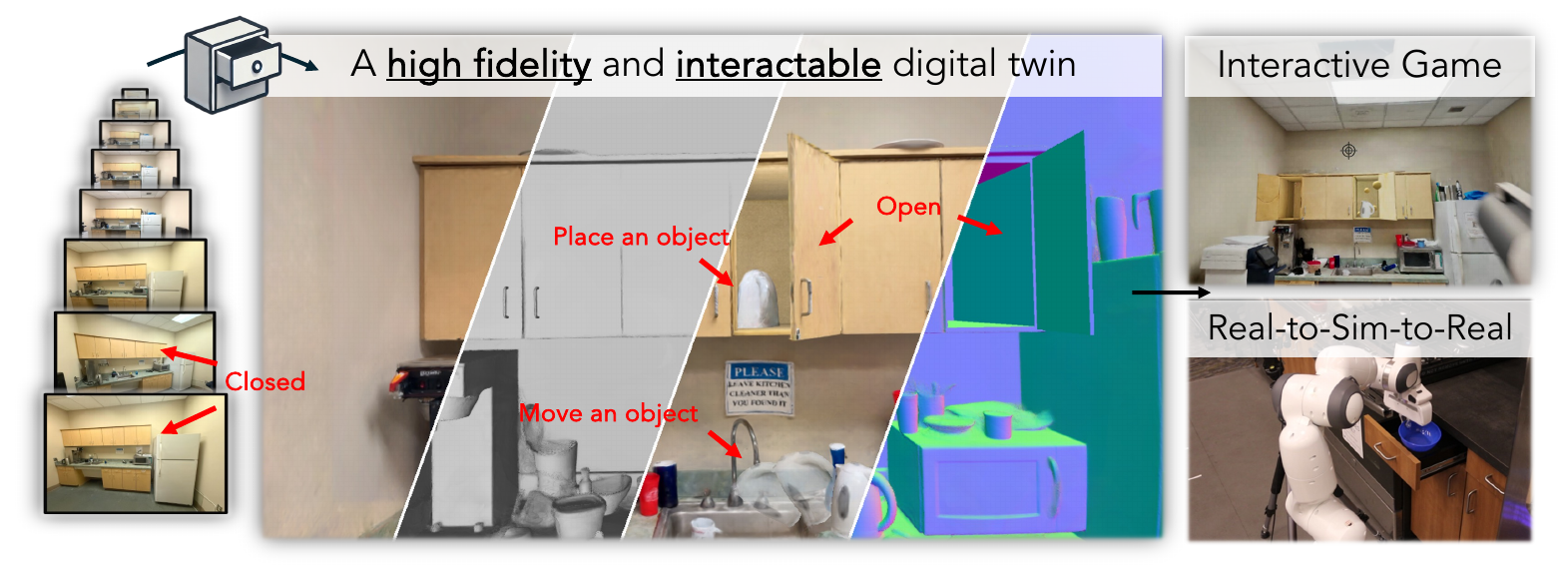}
    \hfill    
\end{center}
\vspace{-9mm}
\captionof{figure}{
\textbf{\modelname} automatically converts a video of a static scene into an \textbf{interactable} and \textbf{actionable} virtual environment.
The reconstructed digital twin features precise geometry, high-fidelity rendering, and supports physical interactions like opening/closing drawers/cabinets and moving/placing objects. 
It can also be seamlessly integrated with modern game engines and robotic simulation platforms, enabling the creation of interactive games and facilitating real-to-sim-to-real policy transfer. 
}
\label{fig:teaser}
}
\vspace{3mm}
]

\input{sec/0_abstract}

\input{sec/1_intro_v3}

\input{sec/2_related_trim}
\input{sec/3_method}

\input{sec/4_exp}

\input{sec/5_conclusion}
{
    \small
    \bibliographystyle{ieeenat_fullname}
    \bibliography{main}
}

\end{document}

%% file: preamble.tex
\usepackage[dvipsnames]{xcolor}

%% file: macros.tex
\newcommand{\todocite}[1]{\textcolor{blue}{Citation needed []}}

\makeatletter
\def\@onedot{\ifx\@let@token.\else.\null\fi\xspace}
\DeclareRobustCommand\onedot{\futurelet\@let@token\@onedot}

\def\etal{\emph{et al}\onedot}

\newcommand\shortcite[2][]{%
  \ifNAT@numbers\cite[#1]{#2}\else\citeyearpar[#1]{#2}\fi}
\newcommand{\ignorethis}[1]{}

\makeatletter
\DeclareRobustCommand\onedot{\futurelet\@let@token\@onedot}
\def\@onedot{\ifx\@let@token.\else.\null\fi\xspace}

\def\etal{\emph{et al}\onedot}
\makeatother

\definecolor{citecolor}{RGB}{34,139,34}
\definecolor{mydarkblue}{rgb}{0,0.08,1}
\definecolor{mydarkgreen}{rgb}{0.02,0.6,0.02}
\definecolor{mydarkred}{rgb}{0.8,0.02,0.02}
\definecolor{mydarkorange}{rgb}{0.40,0.2,0.02}
\definecolor{mypurple}{RGB}{111,0,255}
\definecolor{myred}{rgb}{1.0,0.0,0.0}
\definecolor{mygold}{rgb}{0.75,0.6,0.12}
\definecolor{myblue}{rgb}{0,0.2,0.8}
\definecolor{mydarkgray}{rgb}{0.66,0.66,0.66}

%% file: sec/0_abstract.tex
\begin{abstract}
Creating virtual digital replicas from real-world data unlocks significant potential across domains like gaming and robotics. In this paper, we present \modelname, a novel framework that converts a video of a static indoor scene into a \emph{photorealistic} and \emph{interactive} digital environment. Our approach centers on two main contributions: (i) a reconstruction module based on a \emph{dual scene representation} that reconstructs the scene with \emph{fine-grained geometric details}, 
and (ii) an \emph{articulation} module that identifies articulation types and hinge positions, reconstructs simulatable shapes and appearances and integrates them into the scene. The resulting virtual environment is photorealistic, interactive, and runs in real time, with compatibility for game engines and robotic simulation platforms. We demonstrate the potential of \modelname\ by using it to automatically create an interactive game in Unreal Engine and to enable real-to-sim-to-real transfer for robotics applications. Project page: \href{https://xiahongchi.github.io/DRAWER/}{here}. 
\end{abstract}

%% file: sec/1_intro_v3.tex
\begin{figure*}[t]
    \centering    
    \vspace{-4mm}
    \includegraphics[width=0.98\textwidth,trim={0 0 0 1.6cm}, clip]{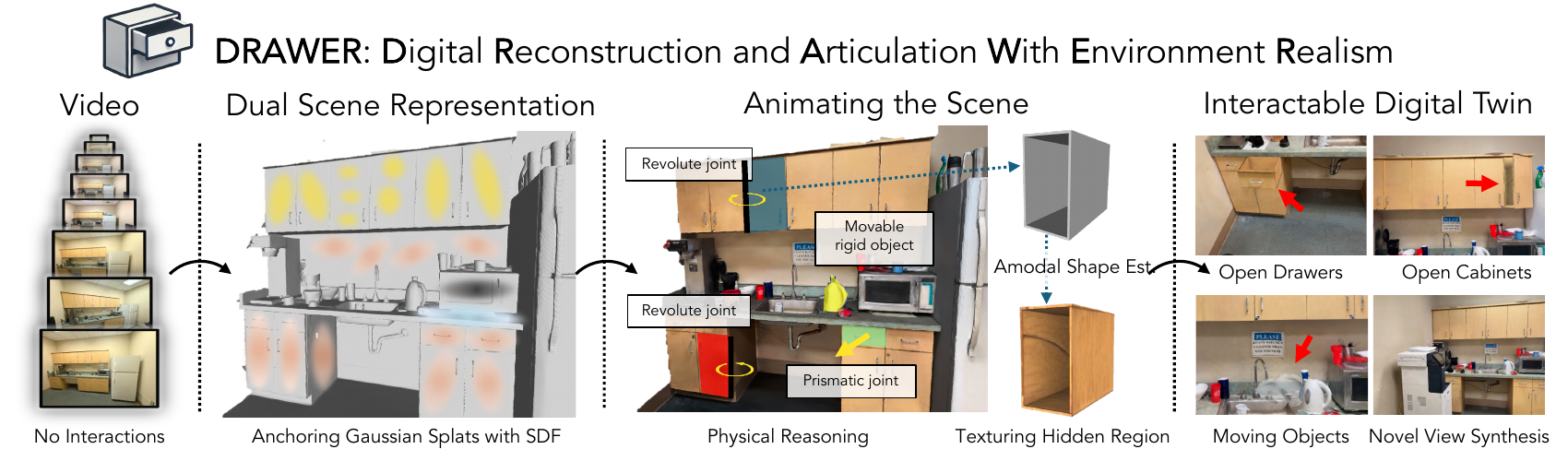}
    \vspace{-3mm}
    \caption{
    \textbf{Overview of \modelname:}  Given multiple posed images from a single video, we first employ a \emph{dual scene representation} to capture high-fidelity visual appearance as well as fine-grained geometry. Then we \emph{animate the scene} by reasoning about articulated and movable rigid-body objects. Finally, our amodal shape estimation with hidden region texturing enables us to create a \emph{complete}  digital twin. Our reconstructions support real-time physical interactions such as opening drawers/cabinets, moving objects, and rendering novel views.
    }
    \label{fig:method}
\end{figure*}

\vspace{-10pt}

\section{Introduction}
\label{sec:intro}
The ability to \emph{automatically} create a \emph{realistic}, \emph{interactable}, and \emph{highly detailed} virtual replica of a physical environment offers immense potential across multiple domains. 
For game developers, this presents an opportunity to replace painstaking human labor with streamlined, automated processes \cite{gta,dosovitskiy2017carla}. 
In robotics, training and evaluating autonomous systems within richly detailed virtual spaces enable safer and more scalable learning. 
Take Fig.~\ref{fig:teaser} as an example. 
Given a video of a scene (in this case, static), if we were to construct a digital replica that is not only visually and geometrically authentic but also physically grounded, then an agent deployed in this mirror world would be able to freely navigate the environment, interact with the scene (\emph{e.g.}, opening drawers/cabinets, grabbing objects), and leverage observations and feedback to learn a policy that can seamlessly transfer to its real-world counterpart. 
Digital twins can thus serve as dynamic, virtual testbeds for studying and interacting with reality. 
However, to this day, automatically generating digital twins that mirror their real-world counterparts in terms of 
\emph{visual appearance}, \emph{geometric details}, and \emph{physical properties} still remains a complex and unresolved task. 

To mitigate the domain gap in visual or geometric quality, 
3D reconstruction techniques, such as neural fields \cite{mildenhall2020nerf,wang2021neus}, have emerged as promising solutions for constructing digital twins. 
However, despite their impressive realism, the reconstructions are still \emph{static} and \emph{non-actionable}. While users can freely view the scene from different angles, they \emph{cannot interact} with it. 
Furthermore, there is often a trade-off between visual fidelity and geometric precision: pushing for photorealism often requires sacrificing some underlying geometric accuracy \cite{mildenhall2020nerf,kerbl20233d}, and vice versa \cite{wang2023neus2,yu2022monosdf}. 
Alternatively, to enable interaction capabilities in digital representations, researchers have utilized shape primitives and CAD models to approximate the physical world \cite{chen2024urdformer,dai2024acdc}. 
While they can construct virtual scenes and objects that resemble their real-world counterparts semantically and functionally, it comes at the expense of visual and geometric fidelity. 
The status quo calls for a method that takes the best of both worlds.

Towards this goal, we present, \modelname, a novel framework that automatically converts a video of a static indoor scene into a \emph{photorealistic} and \emph{interactable} digital environment with \emph{fine-grained geometric details}.   
At the core of our approach lies two key components: (i) a 3D reconstruction module based on a \emph{dual scene representation} and (ii) an \emph{articulation} module. 
Given an input video, we first construct a neural signed distance field (SDF) that effectively captures the scene's geometric details. We then initialize and anchor Gaussian splats with the estimated surface. %
This allows us to avoid floaters and preserve well-behaved geometry while enjoying rendering quality and speed. 
To make the reconstructed scene interactable, 
we leverage foundation models to infer articulation types and hinges of objects in the scene and approximate them with shape primitives. 
To ensure seamless integration of articulated objects into the original reconstruction, we further exploit differentiable rendering to align both their geometry and appearance. 
To enhance realism, we also infer the complete geometry and appearance of hidden interior regions.
The resulting environment is photorealistic, interactable, and runs in real time. 

We evaluate the fidelity of our reconstructed digital twins based on visual realism, articulation accuracy, and the precision of simulated motions across six distinct kitchen environments. \modelname~significantly outperforms prior art across all metrics. To further validate the effectiveness of the generated simulation environments, we employ them to create an interactive game and to train robotic controllers using a real-to-sim-to-real loop. Experimental results indicate that \modelname~eliminates the need for tedious manual effort and hand-specification in both applications.

%% file: sec/2_related_trim.tex
\section{Related Works}
\label{sec:related}

\begin{figure*}[t]
    \centering
    \setlength\tabcolsep{0.05em} %
    \footnotesize
    \vspace{-4mm}
    \resizebox{\textwidth}{!}{
    \begin{tabular}{cccccc}
         \includegraphics[width=0.33\linewidth]{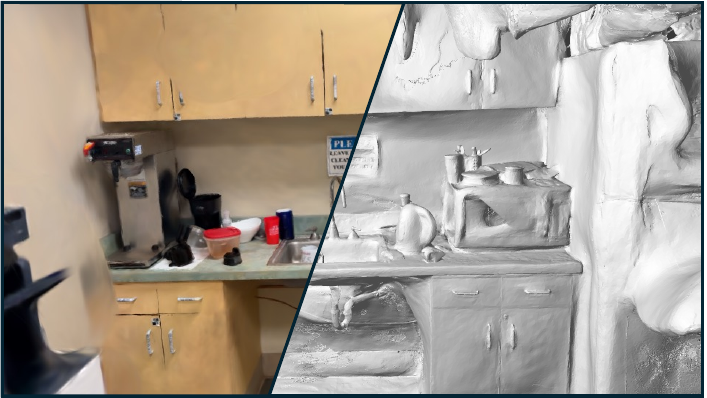} &
        \includegraphics[width=0.33\linewidth]{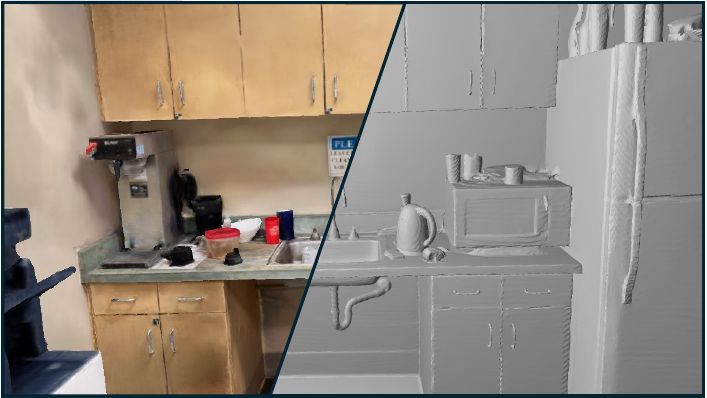} &
        \includegraphics[width=0.33\linewidth]{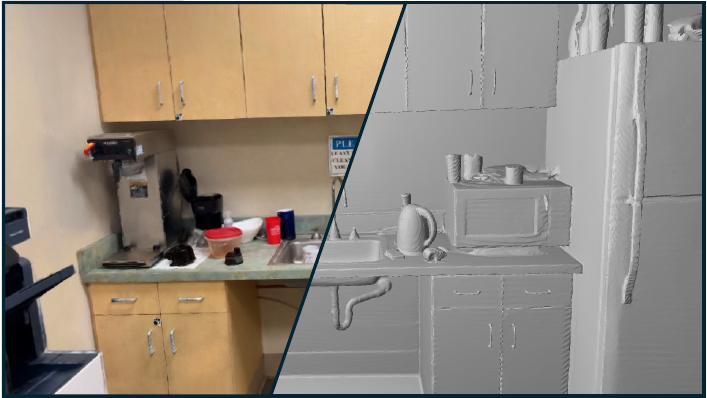} 
        \\        
        Gaussian splatting & Neural SDF & Our Dual Representation \\

    \end{tabular}
    }
    \vspace{-3mm}
    \captionof{figure}{
    \textbf{Qualitative Comparisons of Different Representations:} 
    \textbf{(Left)} Gaussian splatting \cite{kerbl20233d, Huang2DGS2024} can effectively capture the visual appearance of a scene yet struggles with accurate geometric modeling. \textbf{(Middle)} Neural SDF recovers fine-grained geometry but at the cost of slow rendering speeding and degraded appearance modeling.
    \textbf{(Right)} Our dual representation combines the strengths of both representations, offering both high-quality appearance and geometry in real time.    
    }
    \label{fig:compare_dual}
\end{figure*}

\paragraph{Novel view synthesis (NVS):}  The ability to render a scene from new viewpoints using a set of pre-captured images~\cite{chen1993view,levoy1996light,szeliski1998stereo,gortler1998layered,heigl1999plenoptic} is essential for building digital twins. 
The core of NVS is the co-design of scene representations and rendering methods. 
Among common representations such as neural radiance fields~\cite{mildenhall2020nerf, barron2021mip,barron2022mip,mildenhall2022nerf,verbin2022ref}, neural textured mesh~\cite{tang2022nerf2mesh, xia2024video2game}, geometry primitives~\cite{zhao2024surfel, M_ller_2022, yang2020surfelgansynthesizingrealisticsensor, lin2022neurmips, chen2022mobilenerf, aliev2020neuralpointbasedgraphics}, and neural surface fields~\cite{wang2021neus, yariv2021volume, Yu2022SDFStudio, yariv2023bakedsdf, darmon2022improving, oechsle2021unisurf, wu2022voxurf, wang2023neus2, li2023neuralangelo}, Gaussian splatting~\cite{kerbl20233d} emerges as a promising choice, offering flexibility and real-time rendering. On the other hand, neural surface models~\cite{yariv2023bakedsdf} provide accurate and detailed geometry, making them well-suited for reconstruction, articulation, and physical simulation. 
In this paper, we introduce a novel dual representation that combines the best of both works.  Our approach extends beyond standard NVS, enabling active simulation and counterfactual visualization as the scene is interacted with and modified.

\paragraph{Data-driven simulation:} 
Learning-based simulation \cite{manivasagam2020lidarsim,chen2021geosim,amini2022vista,son2022differentiable,yang2023unisim,lu2023urban} has become popular for its effectiveness in simulating dynamics~\cite{, liu2023modelbasedcontrolsparseneural,xie2024physgaussianphysicsintegrated3dgaussians, li2022climatenerf}, modeling lighting \cite{pun2023lightsimneurallightingsimulation, lin2023urbanir}, and generating outputs in response to counterfactual actions \cite{xia2024video2game,liu2024physgenrigidbodyphysicsgroundedimagetovideo,yang2023unisim}.
These methods have been applied in various domains, including content creation~\cite{hsu2024autovfxphysicallyrealisticvideo, liu2023modelbasedcontrolsparseneural}, game development~\cite{xia2024video2game}, robot learning~\cite{yang2020surfelgan,manivasagam2020lidarsim,chen2021geosim, yang2023unisim, pun2024neural,yang2023emernerf}, and multi-modal generation like LiDAR~\cite{manivasagam2020lidarsim,zyrianov2022learning,ultralidar,liu2023real, zyrianov2024lidardmgenerativelidarsimulation, yang2023unisim}. 
Our work falls within this category. We enable realistic modeling, simulation, and rendering of articulated objects in static scenes, from and to photorealistic videos. To our knowledge, this is the first approach of its kind.
The closest work to ours is Video2Game~\cite{xia2024video2game}, which also aims to reconstruct an interactable 3D scene from a video of a static scene. However, there are three key differences: 1) we significantly enhance interactivity by simulating articulated objects; 2) we improve visual quality with a novel dual representation; and 3) beyond real-time gaming, we show utility for robot learning, where robots practice opening drawers and cabinets in our simulated environment and transfer these skills to the real world in a zero-shot setting.

\paragraph{Articulation modeling and simulation:} 
Creating interactable and articulated virtual scenes that resemble reality typically requires specialized skills, professional software, and extensive human efforts~\cite{dosovitskiy2017carla, todorov2012mujoco, mittal2023orbit}. 
To address this challenge, researchers have proposed using automated tools, such as procedural generation~\cite{ProcTHOR,raistrick2024infinigenindoorsphotorealisticindoor}, or learning-based methods to directly model or approximate real-world environments~\cite{jiang2022ditto,liu2023building,hsu2023ditto,yang2023unisim,xia2024video2game}. 
However, current research in robotics and vision primarily focuses on individual objects~\cite{mandi2024real2code,le2024articulate, yi2018deep, hayden2020nonparametric, huang2021multibodysync, noguchi2022watch, lei2023nap,iliash2024s2o,liu2024singapo,chen2024urdformer} and is not directly applicable to larger scenes.
While some methods focus on scene-level modeling, they often assume access to dynamic scenes \emph{before} and \emph{after} interactions \cite{jiang2022ditto,hsu2023ditto,liu2025building}, or rely on human interventions \cite{torne2024reconciling, kerr2024robot}.
To scale this to the real world, creating interactable scenes from \emph{passive observations} has gained increasing attention \cite{chen2024urdformer, dai2024acdc}. 
However, these methods focus on replicating real-world semantics and functionality, often neglecting visual and geometric fidelity. 
In contrast, our digital twins faithfully reproduce real-world environments with high visual, geometric, and physical accuracy -- even without any observed physical interactions during capture. This fidelity is particularly crucial in domains where appearance matters, such as content creation and sim2real applications.
Furthermore, as we will show later, our approach, grounded in precise geometry, achieves superior accuracy in articulation reasoning compared to prior methods.

\paragraph{Controllable video generation:} 
An alternative to modeling how our world works is to leverage video generative models to (implicitly) simulate various effects \cite{klingai,opensora,yang2023learning,li2024generative,zhang2024physdreamer,hao2018controllable,zhang2023controlvideo,guo2023animatediff,wang2024videocomposer}. 
While existing methods produce promising image space dynamics, they lack access to internal states, which are essential for tasks like mobile manipulation. 
For instance, knowing if a robot has grasped an object or opened a drawer is critical. 
Additionally, generated frames often degrade in quality over longer time spans, and integrating video dynamics with physical models or simulation engines remains challenging. In contrast, our approach adheres to physical laws, is compatible with simulation engines, and provides access to underlying states beyond visual rendering.

\begin{figure}[t]
    \centering    \setlength\tabcolsep{0.05em} %
    \scriptsize   
    \vspace{-2mm}
    \resizebox{0.99\linewidth}{!}{
    \begin{tabular}{cc}
        \includegraphics[width=0.45\linewidth]{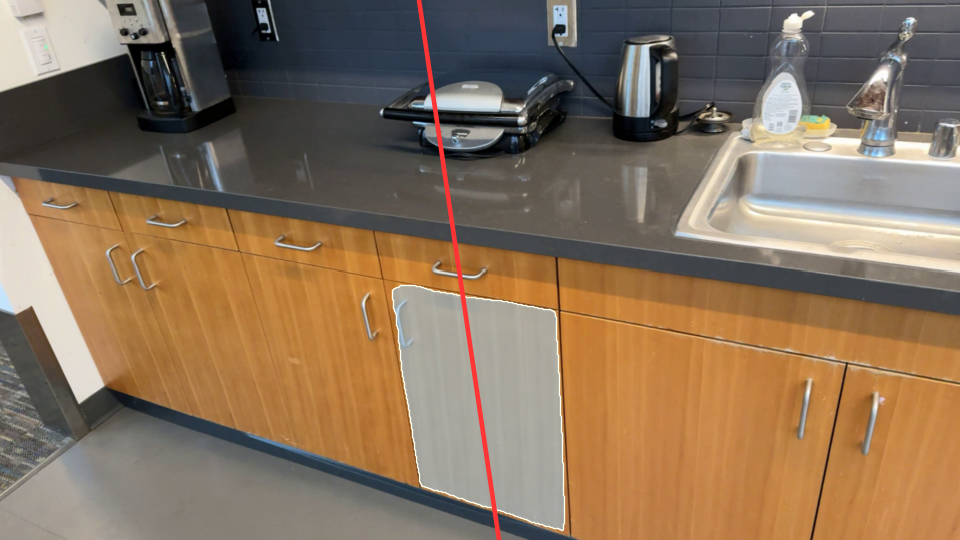} & 
        \includegraphics[width=0.45\linewidth]{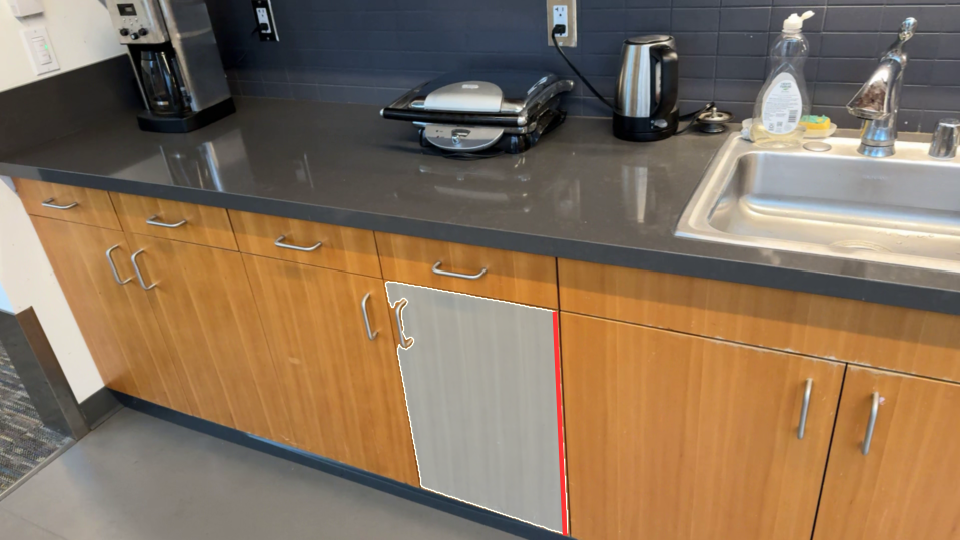} 
        \\
         3DOI \cite{qian2023understanding} & Ours 
    \end{tabular}
    }
    \vspace{-3mm}
    \caption{
    \textbf{Articulation Estimation:} We visualize the estimated revolute axes and articulated object masks produced by 3DOI \cite{qian2023understanding} and {\modelname}, demonstrating that {\modelname} achieves more precise articulation estimation due to its underlying 3D geometry.
    }
    \label{fig:art-est}
\end{figure}

%% file: sec/3_method.tex
\begin{table*}[t]
\centering
\resizebox{0.95\textwidth}{!}{
    \begin{tabular}{lcccccccc}
       \toprule
       {\multirow{2}{*}{\textbf{Method}}} & \multirow{2}{*}{\textbf{Representation}} & \multirow{2}{*}{\textbf{PSNR}$\uparrow$} & \multirow{2}{*}{\textbf{SSIM}$\uparrow$}  & \multirow{2}{*}{\textbf{LPIPS}$\downarrow$}  & \multicolumn{4}{c}{\textbf{Interactive Compatibility}}  \\
       & & &&&\textbf{Real time} &  \textbf{Rigid-body physics} & \textbf{Scene decomposition} & \textbf{Articulation} \\
        \midrule
        Nerfacto \cite{tancik2023nerfstudio} & Volume & 25.49 & 0.911 &  0.163 & \xmark & \xmark & \xmark & \xmark\\
        Video2Game (NeRF) \cite{xia2024video2game} & Volume & 27.95 & 0.884 &  0.239 & \xmark & \xmark & \xmark & \xmark\\
        \arrayrulecolor{lightgray}\midrule\arrayrulecolor{black}
        BakedSDF* \cite{yariv2023bakedsdf} & Mesh & 21.11 & 0.787 &  0.409 & \cmark & \cmark & \xmark & \xmark\\
        Video2Game (Mesh) \cite{xia2024video2game} &  Mesh & 22.63 & 0.822 & 0.323  & \cmark & \cmark & \cmark &  \xmark \\
        \arrayrulecolor{lightgray}\midrule\arrayrulecolor{black}
        3DGS \cite{kerbl20233d} & Points & 30.42 & 0.954 & 0.126 & \cmark & \xmark & \xmark & \xmark  \\        
        2DGS* \cite{Huang2DGS2024} & Points & 25.68 & 0.884 & 0.269  & \cmark & \cmark & \xmark & \xmark \\
        \arrayrulecolor{lightgray}\midrule\arrayrulecolor{black}
        Ours & Points+Mesh & {27.80} &  {0.912} & {0.159} & \cmark & \cmark & \cmark & \cmark \\
        \bottomrule
    \end{tabular}
    }
    \vspace{-2mm}
    \caption{
    \textbf{Quantitative Results on Novel View Synthesis and Interactive Compatibility Analysis:} The dual scene representation in {\modelname} provides competitive high-fidelity rendering with most-capable interactive compatibility. {$^*$BakedSDF \cite{yariv2023bakedsdf} and 2D-GS \cite{Huang2DGS2024} represent the entire scene as a whole, limiting object-level interactions.}
    }
    \label{tab:nvs}
\end{table*}

\section{DRAWER: Digital Reconstruction and Articulation With Environment Realism}
\label{sec:method}

Given a video of a static scene, our goal is to develop an \emph{interactable} and \emph{actionable} digital twin that replicates the 3D world \emph{geometrically}, \emph{photometrically}, \emph{physically},
and \emph{efficiently}. 
Based on the observation that existing approaches tend to either focus on appearance modeling while neglecting physical interaction \cite{mildenhall2020nerf}, or prioritize interaction at the expense of realism \cite{chen2024urdformer},
we carefully design our method to fulfill all essential properties needed for realistic, real-time interactive applications.  
At the core of our approach is a \emph{compositional dual scene representation} that effectively and efficiently supports both sensor and physics simulations.
By decomposing the world into individual entities and modeling them with diverse yet tightly coupled representations,
we can capture various modalities (\emph{e.g.}, RGB, depth) and enable physical interactions without compromising fidelity. 
Fig. \ref{fig:method} shows an overview of our approach.

\subsection{Preliminaries}
\label{sec:prelim}
\paragraph{Neural signed distance fields (SDFs):} 
A neural SDF $f_\theta^\text{SDF}$ maps
a 3D point $\mathbf{x}\in\mathbb{R}^3$ and a view direction $\mathbf{d}\in\mathbb{R}^2$ to an RGB radiance $\mathbf{c}\in\mathbb{R}^3$ and a {signed} distance to the nearest surface $s \in \mathbb{R}$: $s, \mathbf{c} = f_\theta^\text{SDF}(\mathbf{x}, \mathbf{d})$. 
One popular paradigm to learn neural SDF from a set of posed images is through volume rendering \cite{max1995optical,mildenhall2021nerf,yariv2021volume}. By converting signed distances to volume densities \cite{wang2021neus,yariv2023bakedsdf}, one can alpha-composite the %
radiance of 3D points $\mathbf{c}_i$ along each camera ray $\mathbf{r}$ to obtain the estimated pixel color $\mathbf{c}(\mathbf{r}) = \sum_{i=1}^{N}w_i\mathbf{c_i}$ , and then compare with the GT: $\mathcal{L}_\text{rgb} = \sum_{\mathbf{r}}\lVert \hat{\mathbf{c}}(\mathbf{r}) -  \mathbf{c}(\mathbf{r})\rVert_2^2$. 
Here, $w_i = \alpha_i \prod_{j=1}^{i-1} (1-\alpha_j)$ indicates  blending weight, and $\alpha_i$ represents opacity. 
We refer the readers to \cite{yariv2021volume, yariv2023bakedsdf}  on how to derive opacity from signed distances.
In practice, although Neural SDFs are better at capturing geometry, their rendering quality often lags behind NeRF-based approaches \cite{mildenhall2021nerf,barron2021mip}. Also, 
volume rendering requires sampling many points per ray, making it time-consuming and unsuitable for high-FPS applications. 
To improve efficiency, one strategy is to convert Neural SDFs into meshes. While this significantly accelerates rendering, it compromises rendering quality 
 \cite{xia2024video2game}

\paragraph{Gaussian splatting:} An alternative approach for maintaining high visual quality with efficiency is to represent the scene as a set of 3D Gaussians \cite{kerbl20233d}. Each Gaussian is characterized by a set of parameters: mean $\bm{\mu}$, scale $\mathbf{S}$, rotation $\mathbf{R}$, opacity $\alpha$, and color radiance $\mathbf{c}$ (encoded using spherical harmonics). The covariance is derived as $\mathbf{\Sigma} = \mathbf{R}\mathbf{S}\mathbf{S}^T\mathbf{R}^T$. 
By rasterizing the 3D Gaussians and alpha-compositing them, we obtain the pixel color $\mathbf{c}(\mathbf{p}) = \sum_{i=1}^N w_i \mathbf{c}_i$. 
We optimize all parameters by minimizing the photometric error $\mathcal{L}_\text{rgb} = \sum_{\mathbf{p}}\lVert \hat{\mathbf{c}}(\mathbf{p}) -  \mathbf{c}(\mathbf{p})\rVert_2^2$. 
A key property of Gaussian splatting is its support for adaptive density control. By dynamically spawning new Gaussians and culling redundant ones, the method effectively synthesizes both low- and high-frequency details.
Gaussian splats are also inherently composible and controllable \cite{luiten2023dynamic, xie2024physgaussianphysicsintegrated3dgaussians}. 
While Gaussian splatting enables real-time, photorealistic rendering of complex scenes, the underlying geometry can be unsatisfactory. 
In practice, there are often `floating' Gaussians in free space {misaligned with the underlying geometry} \cite{, kerbl20233d,guédon2023sugarsurfacealignedgaussiansplatting, yu2024gaussian}.

\begin{table}[t]
    \centering
    \setlength\tabcolsep{0.05em} %
    \resizebox{0.48\textwidth}{!}{
    \begin{tabular}{c}
          \includegraphics[trim={5.2cm 0 0 1cm},clip]{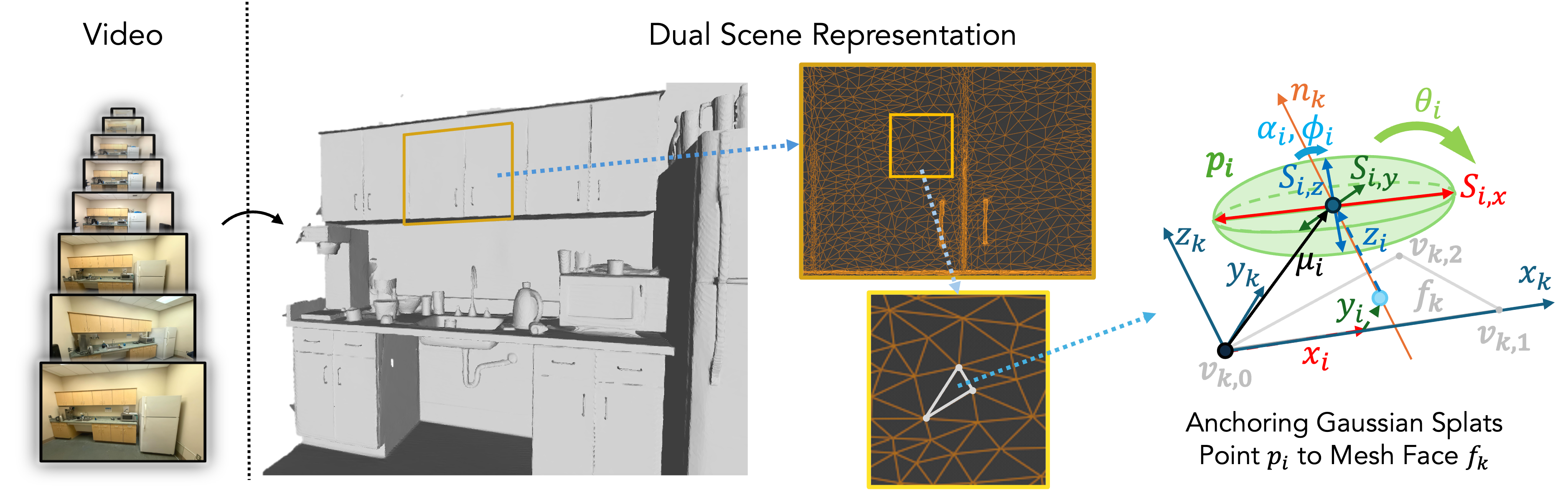} 
    \end{tabular}
    }
    \vspace{-2mm}
    \captionof{figure}{Our \textbf{dual scene representation} combines Neural SDF and Gaussian splatting. We anchor Gaussians around the reconstructed mesh (zero-level set) extracted from the SDF. For details on our Gaussian splat parameterization, please refer to the supp. material.
    }
    \label{tab:dual_diagram}
\end{table}

\setlength{\fboxsep}{0pt}
\setlength{\fboxrule}{0.5pt}
\begin{figure*}[t]
    \centering
    \setlength\tabcolsep{0.05em} %
    \footnotesize
    \vspace{-4mm}
    \resizebox{\textwidth}{!}{
    \begin{tabular}{cccc}

        \raisebox{39px}{\fbox{{\begin{varwidth}{\dimexpr\textwidth-2\fboxsep-2\fboxrule\relax}
        \includegraphics[width=0.24\linewidth]{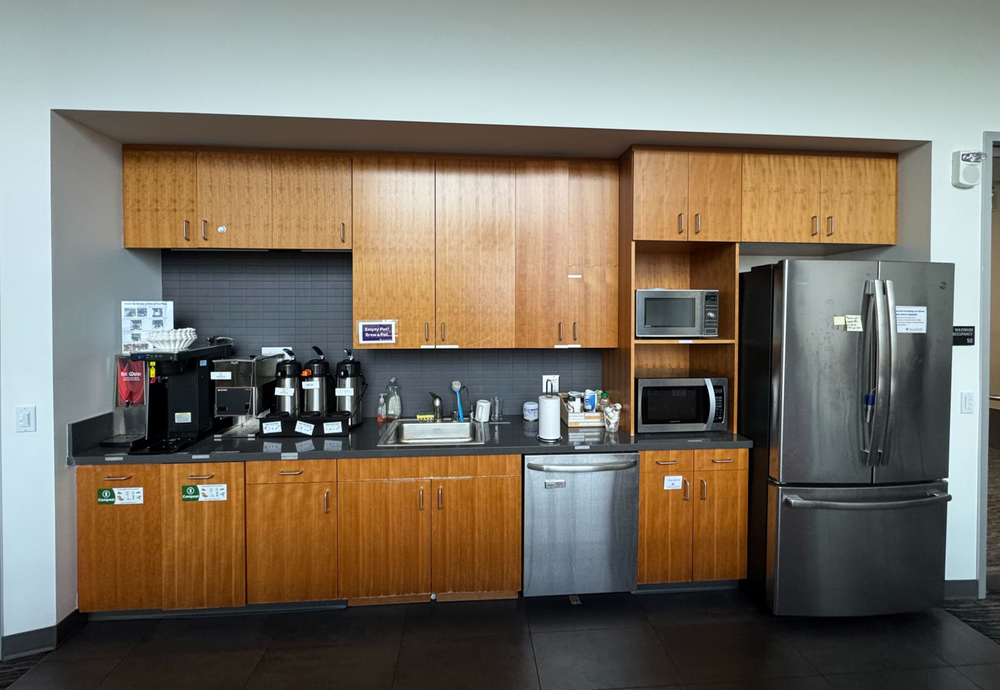}
        \end{varwidth}}}} & 
        \raisebox{39px}{\fbox{{\begin{varwidth}{\dimexpr\textwidth-2\fboxsep-2\fboxrule\relax}
        \includegraphics[width=0.24\linewidth]{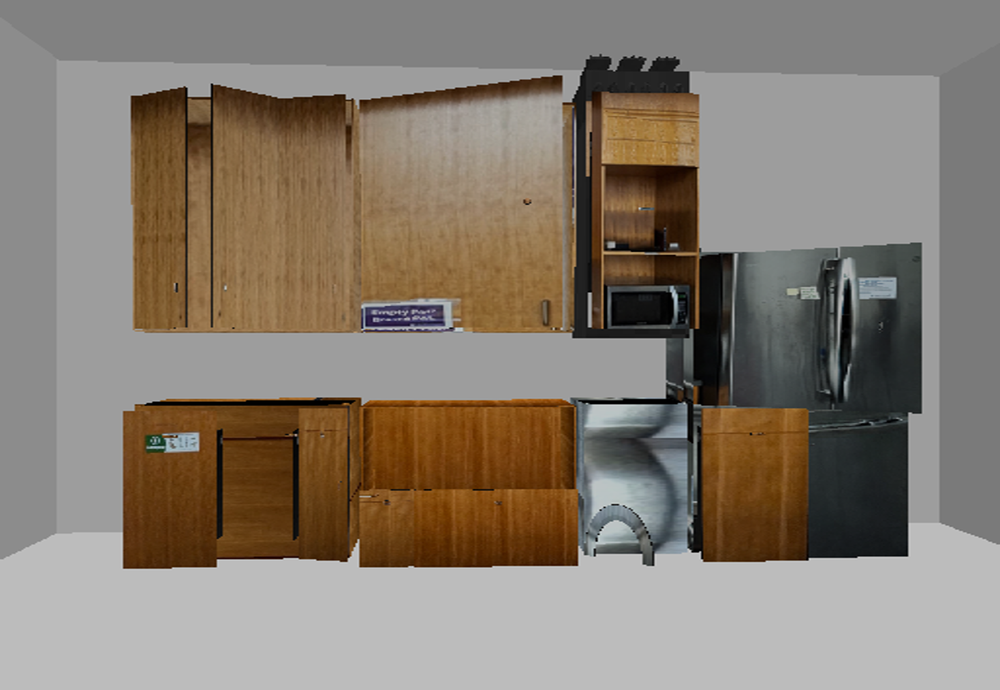} 
        \end{varwidth}}}} & 
        \raisebox{39px}{\fbox{{\begin{varwidth}{\dimexpr\textwidth-2\fboxsep-2\fboxrule\relax}
        \includegraphics[width=0.24\linewidth]{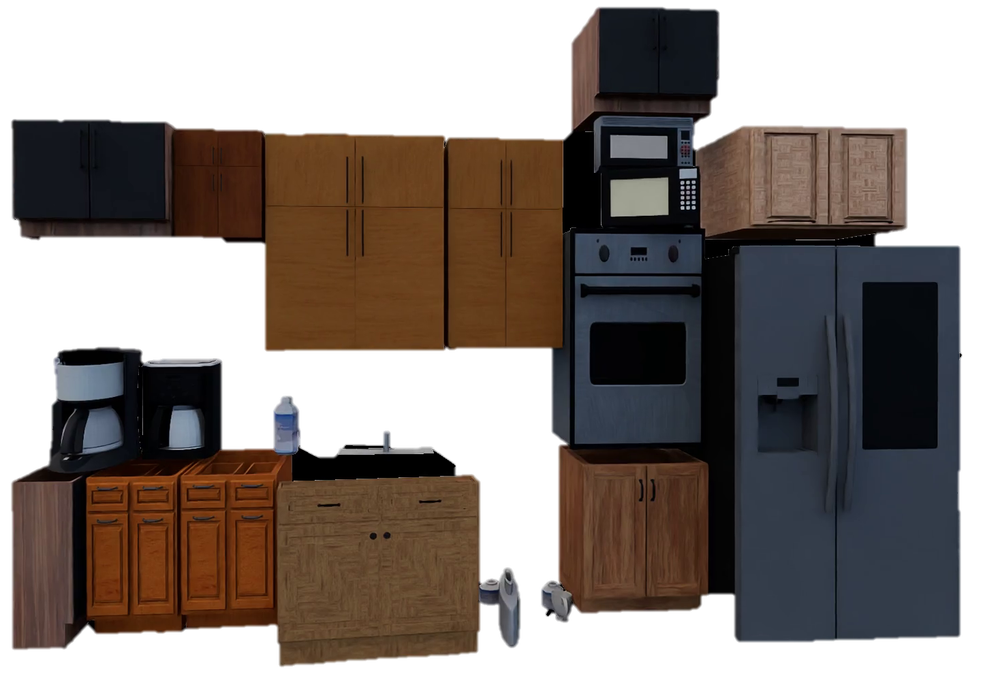} 
        \end{varwidth}}}} & 
        \raisebox{39px}{\fbox{\begin{varwidth}{\dimexpr\textwidth-2\fboxsep-2\fboxrule\relax}
         \animategraphics[autoplay,loop,width=0.295\linewidth, trim={0 0 0 0}, clip]{20}{figures/outvideo/4/}{001}{047}
         \end{varwidth}}}  
         
         \\
         {Input} & URDFormer* \cite{chen2024urdformer} & Digital Cousin* \cite{dai2024acdc} & 
         DRAWER (Ours)
       
    \end{tabular}
    }
    \vspace{-3mm}
    \captionof{figure}{
    \textbf{Qualitative Comparisons on Interactable 3D Reconstruction:} 
    For URDFormer \cite{chen2024urdformer} and Digital Cousin \cite{dai2024acdc}, we select the best results from multiple image runs. Despite these advantages, prior methods still suffer from limited realism and spatial misalignment. In contrast, DRAWER produces reconstructions that are significantly more realistic and faithful.  {To view this figure as a \textbf{video}, we recommend using Adobe Acrobat.} 
    For additional comparisons, please refer to the supp. material.    
    }
    \label{fig:qual-digital-twin}
\end{figure*}

\begin{figure*}[t]
    \centering
    \setlength\tabcolsep{0.05em} %
    \footnotesize
    \resizebox{0.98\textwidth}{!}{
    \begin{tabular}{ccc}
        \includegraphics[width=0.33\textwidth]{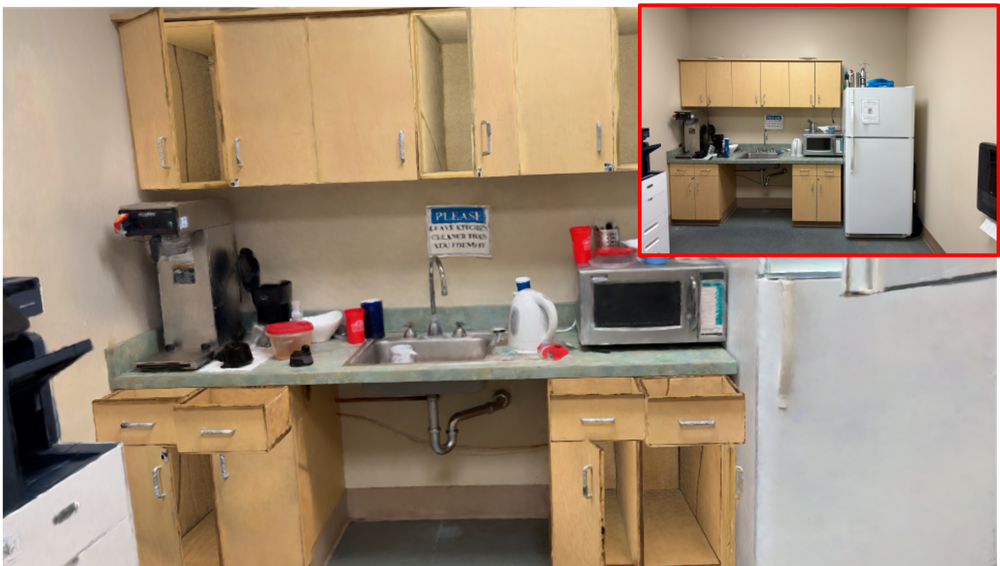} &
        \includegraphics[width=0.33\textwidth]{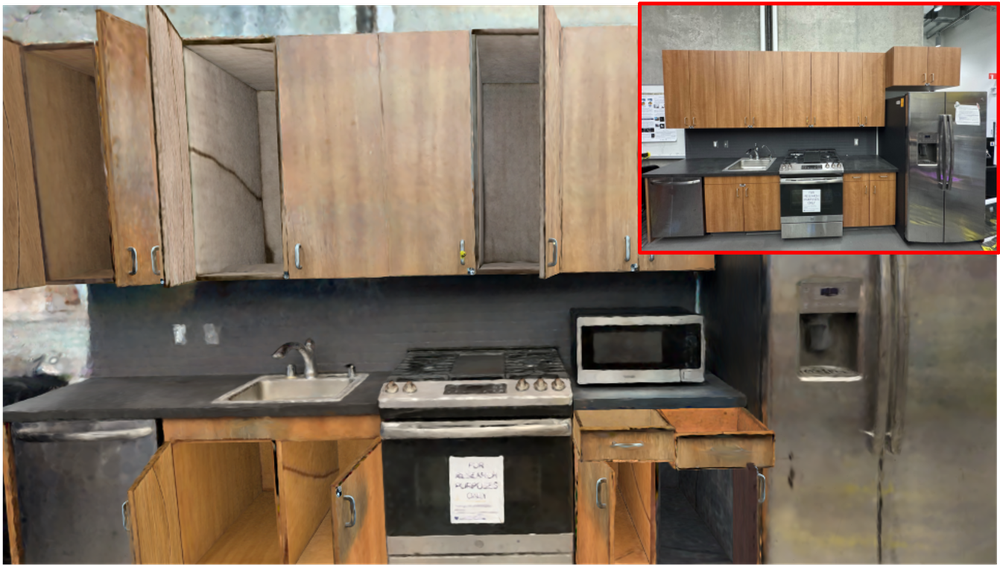}  &
        \includegraphics[width=0.33\textwidth]{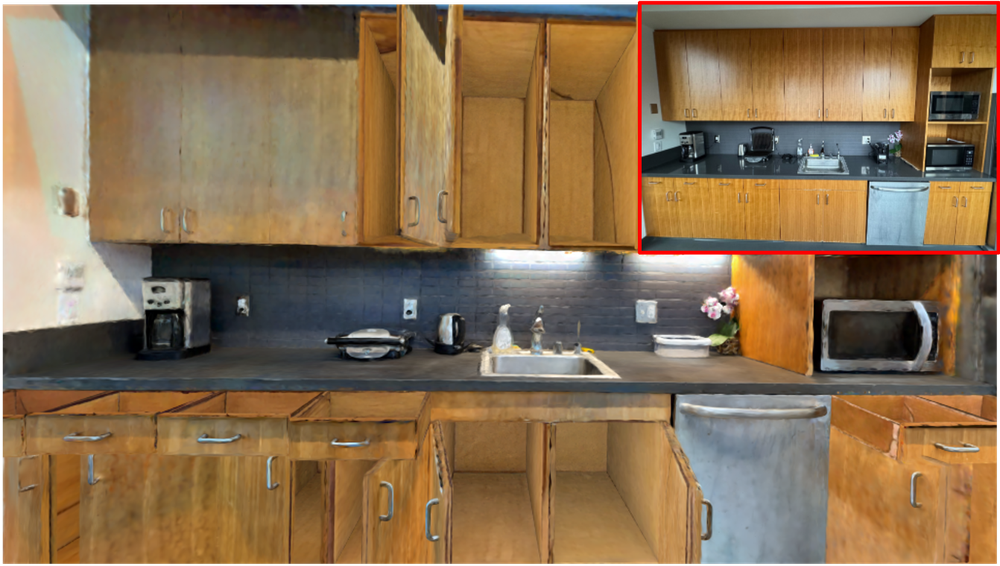}
       
    \end{tabular}
    }
    \vspace{-3mm}
    \captionof{figure}{
    \textbf{Qualitative Results on Additional Kitchen Scenes:} Inset figures display the input/ground truth static observations.
    }
    \label{fig:more_results_kitchens}
\end{figure*}

\subsection{Dual Scene Representation}
\label{sec:dual}
Each 3D representation has its own advantages and limitations, often involving a trade-off among visual fidelity, geometric precision, and speed. To address these inherent constraints, we propose leveraging different representations to capture each individual aspect while enforcing tight coupling between them. 

\paragraph{Geometry:} 
Accurate surface modeling is essential for physical simulation, such as collision {modeling and object manipulation.} 
To capture fine-grained geometric details, we first follow Yariv \etal \cite{yariv2023bakedsdf} and parameterize the scene with a neural SDF. 
Since learning purely from RGB often leads to ambiguities \cite{yu2022monosdf,xia2024video2game}, 
we further leverage off-the-shelf 2D foundation models \cite{garcia2024fine,bae2024dsine} to predict surface normal and depth as regularization.  
We volume render the scene's color, depth, and normal and learn the SDF by jointly optimizing all the losses. 
Please refer to the supp. material for details.

\paragraph{Appearance:} 
High visual quality can significantly enhance immersive experiences in gaming and is essential for sim-to-real visual policy learning. We adopt {3D} Gaussian splatting \cite{kerbl20233d} due to its exceptional efficiency and ability to capture nuanced elements. 
Besides photometric error, we render Gaussian depth maps and use SDF depth rendering to regularize the Gaussians. 

\paragraph{Coupling:} 
One straightforward approach is to align the coordinates of Gaussian splatting with neural SDF, using the former for RGB rendering and the latter for collision {modeling}. However, this can lead to significant mismatch between visual observations and the actual geometry. For example, floating Gaussians may produce visual artifacts when viewed from different angles, creating the illusion of an object in free space
\cite{hsu2024autovfxphysicallyrealisticvideo, xie2024physgaussianphysicsintegrated3dgaussians, guédon2023sugarsurfacealignedgaussiansplatting}. This inconsistency is suboptimal for downstream applications. 

To address this issue, we propose anchoring the Gaussians around the zero level-set of the neural SDF. This approach offers two main advantages: First, it allows the Gaussians to retain some flexibility in movement while avoiding the aforementioned issue. Second, if the scene is interacted with and the underlying scene SDF changes, anchoring the Gaussians to the SDF ensures that appearance changes are automatically handled. 
Since repeatedly querying the learned SDF is computationally expensive, we in practice extract a {high-resolution} mesh from the SDF and anchor Gaussians to it. Specifically, we spawn Gaussians at the centroid of each face. 
The scales $\mathbf{S}$ are initialized to the respective face inradii, and the rotations $\mathbf{R}$ are aligned with the face normals. 
These Gaussians can move freely within the face and a limited distance along the normal direction.
They can also tilt around the normal direction. 
To better align Gaussians with the underlying geometry, we {regularize} the scale along normal directions. 
For adaptive density control, during splitting, 
we ensure that new Gaussians remain on the same face and close to the existing one. We divide the scale by $1.6$ and copy the rest of the remaining parameters. 
We show an illustration in Fig. \ref{tab:dual_diagram} and refer the readers to the supp. material for details.

\paragraph{Learning with straight-through estimator:} 
Restricting each Gaussian to lie within a certain range of its corresponding face is a non-differentiable operation, which makes naive training of Gaussian splatting ineffective (see Sec. \ref{sec:exp}).
To address this, we reparameterize all forward operations that involve \texttt{clip} from $x^\text{o} = \texttt{clip}(x^\text{in})$ to $x^\text{o} = \texttt{sg}(\texttt{clip}(x^\text{in})) + x^\text{in} - \texttt{sg}(x^\text{in})$, where $\texttt{sg}(\cdot)$ denotes stop gradient. We then apply straight-through  estimator \cite{bengio2013estimating} to transfer gradients from after clipping to before clipping.

\paragraph{Relationship to existing work:} 
Our approach is closely related to recent work that combines 3D Gaussian splats with meshes \cite{qian2024gaussianavatars,gao2024mesh,wen2024gomavatar, hsu2024autovfxphysicallyrealisticvideo}. 
However, there exist several key differences.  
First, previous work either \emph{fixes} the positions of Gaussians to the centroids of faces \cite{paudel2024ihuman,wen2024gomavatar} or employs a position loss to \emph{encourage} Gaussians to remain close to the mesh. The former sacrifices flexibility, while the latter cannot guarantee effective coupling. In contrast, we allow Gaussians to move freely within the face and a limited distance along the normal direction, ensuring a balance between flexibility and geometric binding. 
Second, prior work subdivides meshes a priori to obtain denser Gaussian placements \cite{gao2024mesh}, whereas we use adaptive density control to automatically spawn new Gaussians as needed. 
As we will show in Sec. \ref{sec:exp}, these design differences have a significant impact on the final rendering quality.

\begin{figure*}[t]
    \centering
    \vspace{-4mm}
    \setlength\tabcolsep{0.05em} %
    \footnotesize
    \resizebox{\textwidth}{!}{
    \begin{tabular}{ccccc}

        \includegraphics[width=0.2\textwidth]{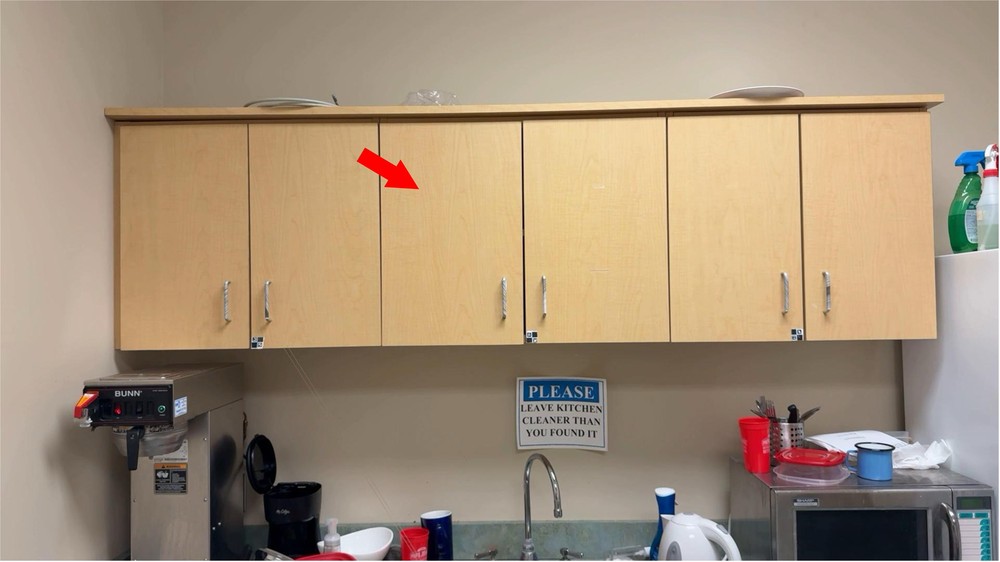} &
        \includegraphics[width=0.2\textwidth,height=0.5625\dimexpr0.2\textwidth\relax,keepaspectratio=false]{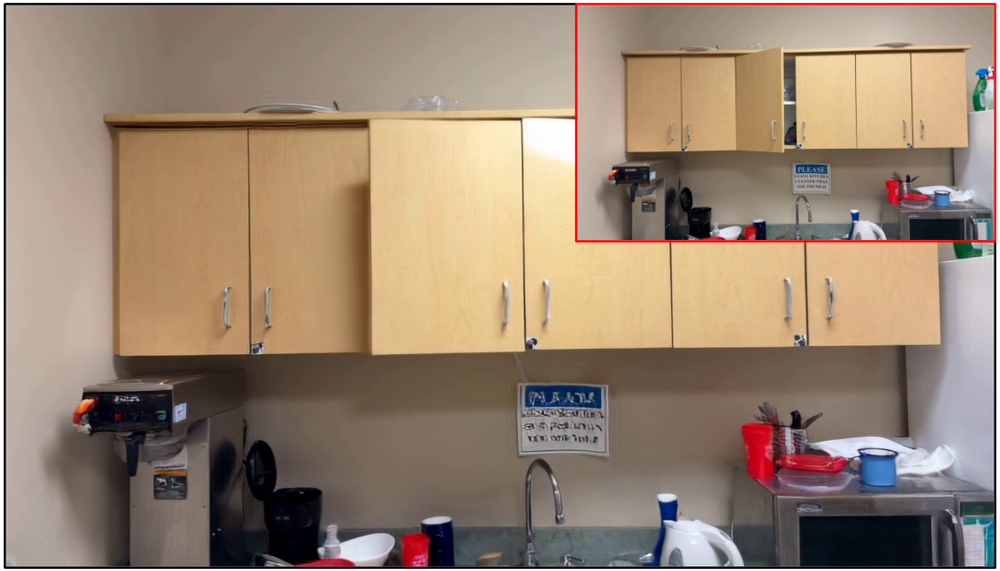} &        
        \includegraphics[width=0.2\textwidth,height=0.5625\dimexpr0.2\textwidth\relax,keepaspectratio=false]{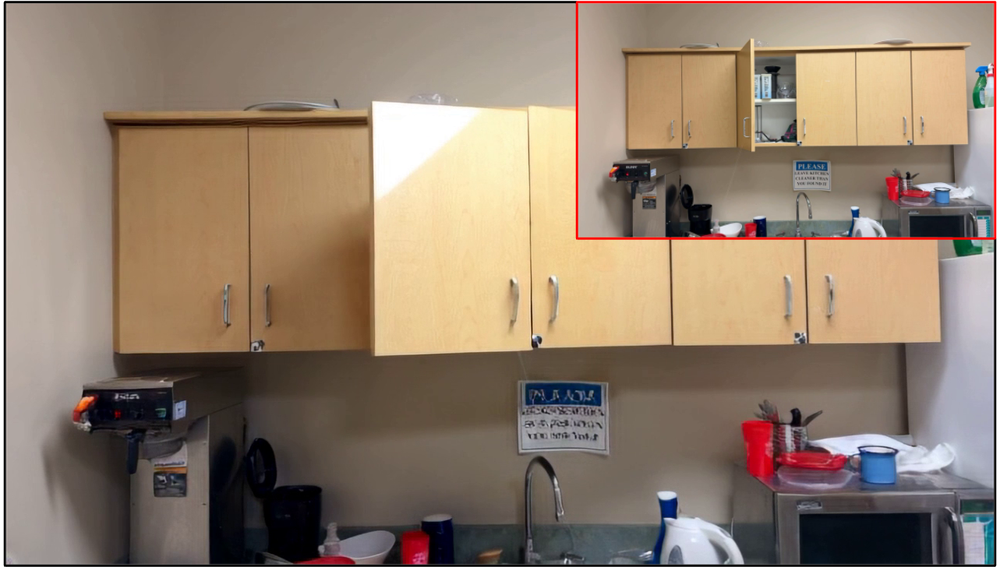} &
        \includegraphics[width=0.2\textwidth,height=0.5625\dimexpr0.2\textwidth\relax,keepaspectratio=false]{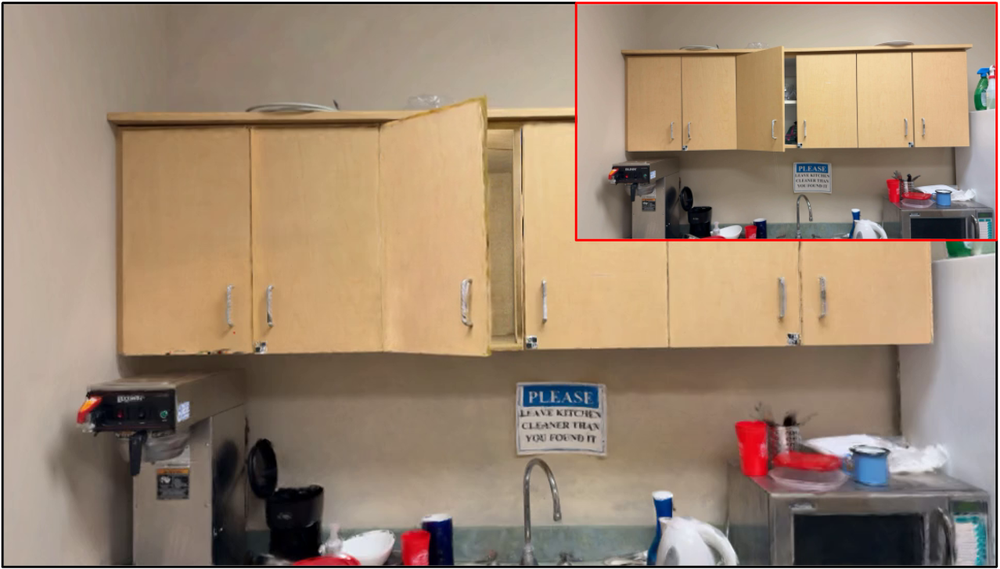} &
        \includegraphics[width=0.2\textwidth,height=0.5625\dimexpr0.2\textwidth\relax,keepaspectratio=false]{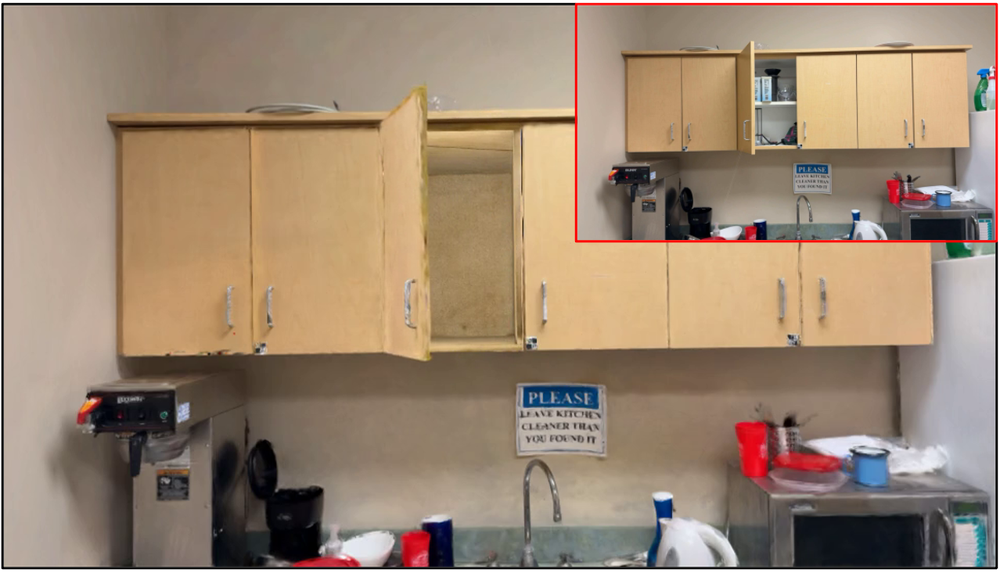} 
         \\

        \includegraphics[width=0.2\textwidth]{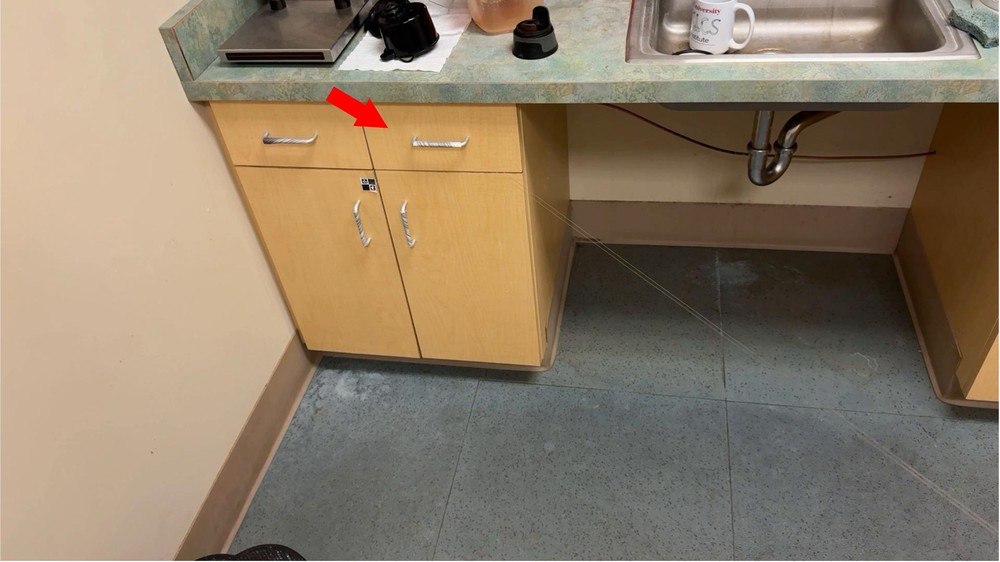} &
        \includegraphics[width=0.2\textwidth]{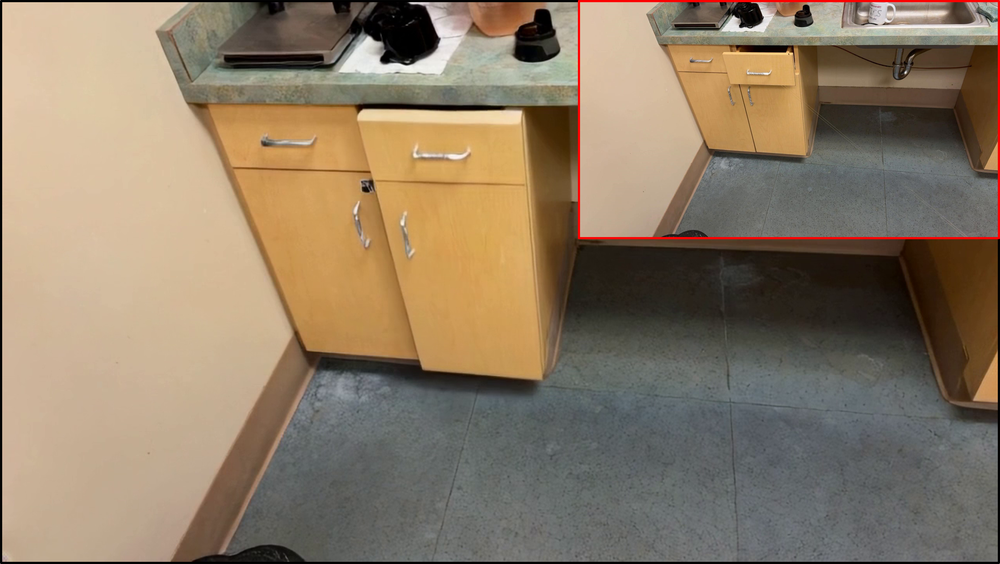} &
        \includegraphics[width=0.2\textwidth]{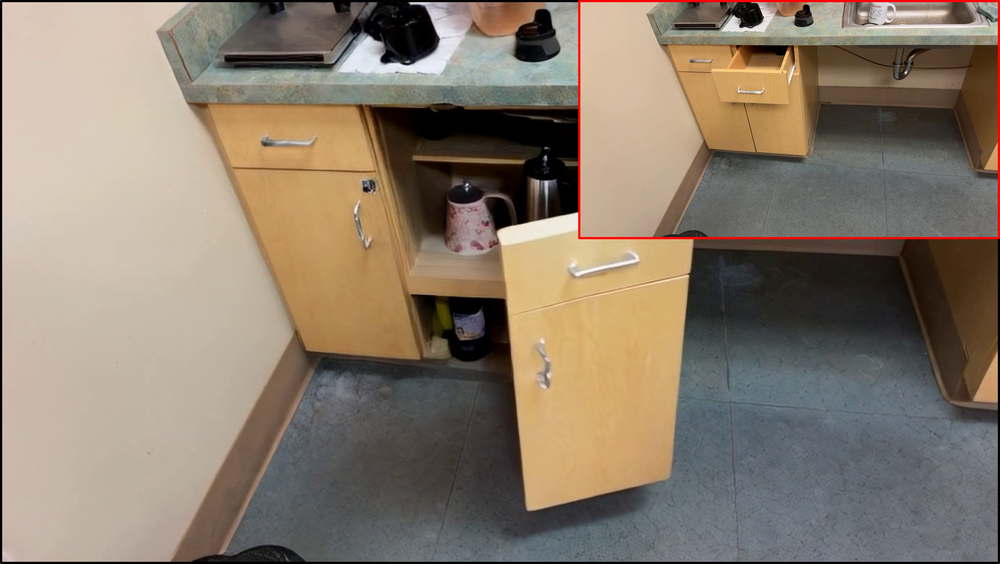} &        
        \includegraphics[width=0.2\textwidth]{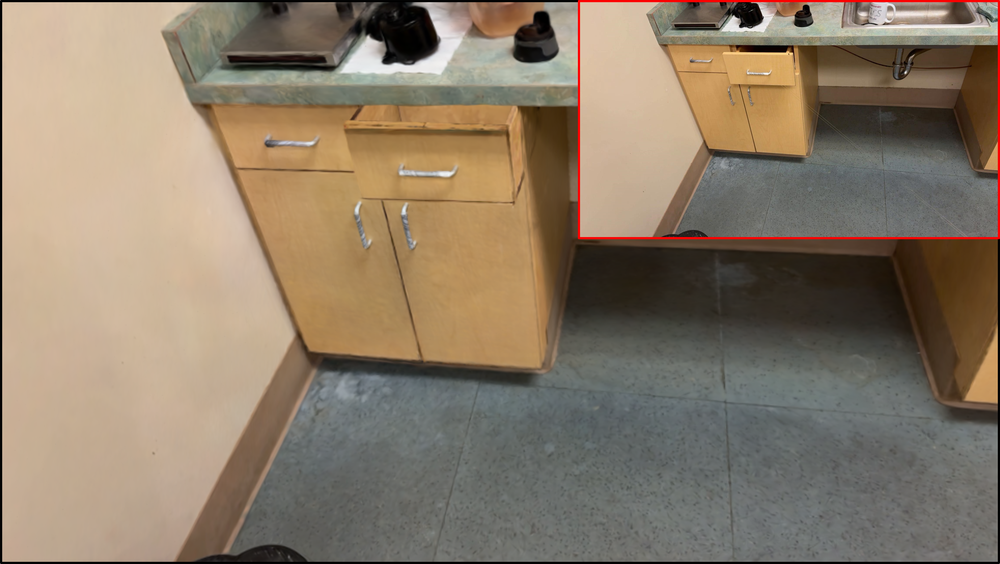} &
        \includegraphics[width=0.2\textwidth]{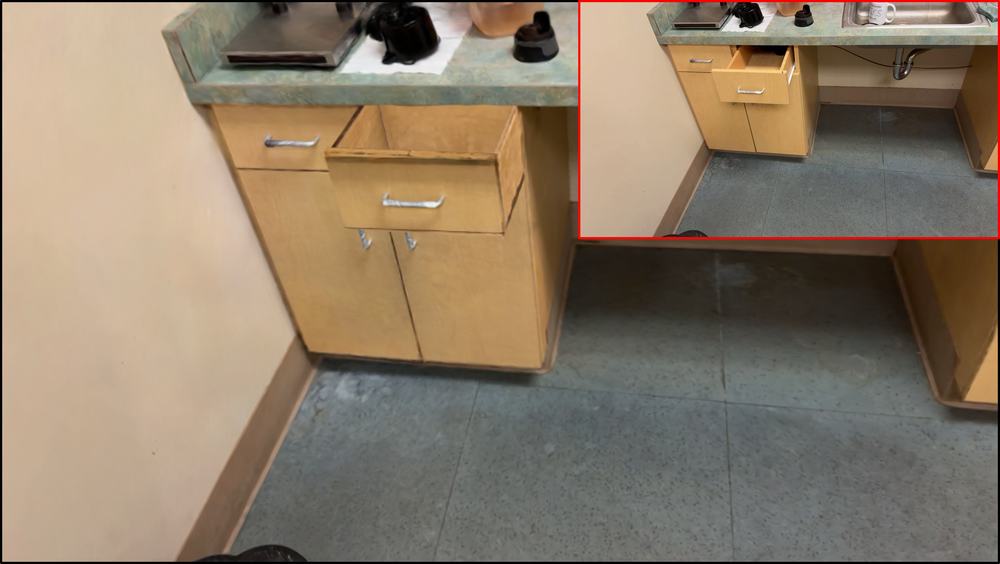} \\

        Input: Closed Cabinet/Drawer & KlingAI: Half Open & KlingAI: Fully Open  & Our Simulation: Half Open & Our Simulation: Fully Open        
    \end{tabular}
    }
    \vspace{-3mm}
    \caption{
    \textbf{Qualitative Comparison on Articulation Simulation:} 
    Our method produces realistic and accurate articulations, whereas KlingAI fails to do so -- even with manual segmentation masks and motion as inputs. Ground truth interactions are shown in the inset figures.
    }
    \label{fig:qual-animate}
\end{figure*}

\begin{figure*}[t]
    \centering
    \setlength\tabcolsep{0.05em} %
    \footnotesize
    \resizebox{\textwidth}{!}{
    \begin{tabular}{cccc}
    
        \includegraphics[width=0.24\linewidth]{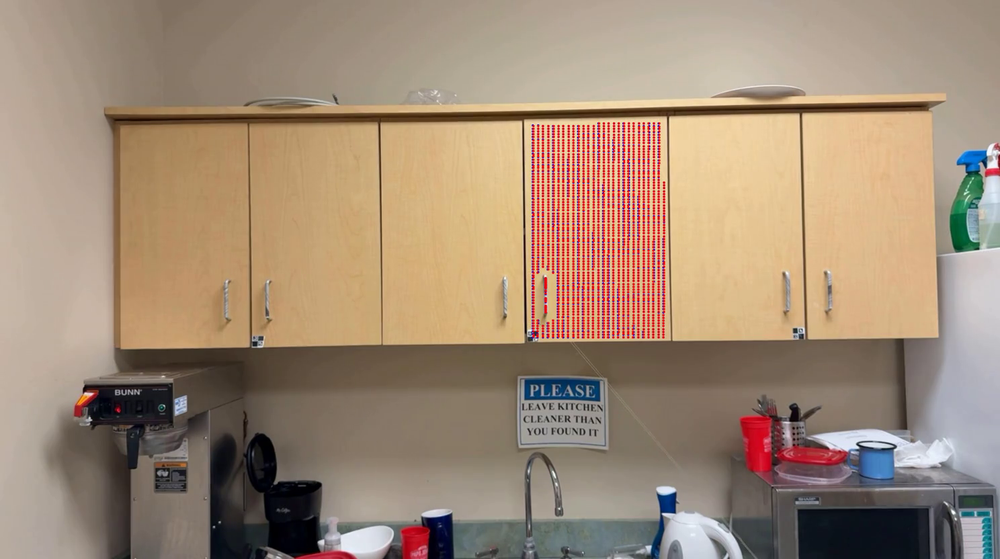} &
        \includegraphics[width=0.24\linewidth]{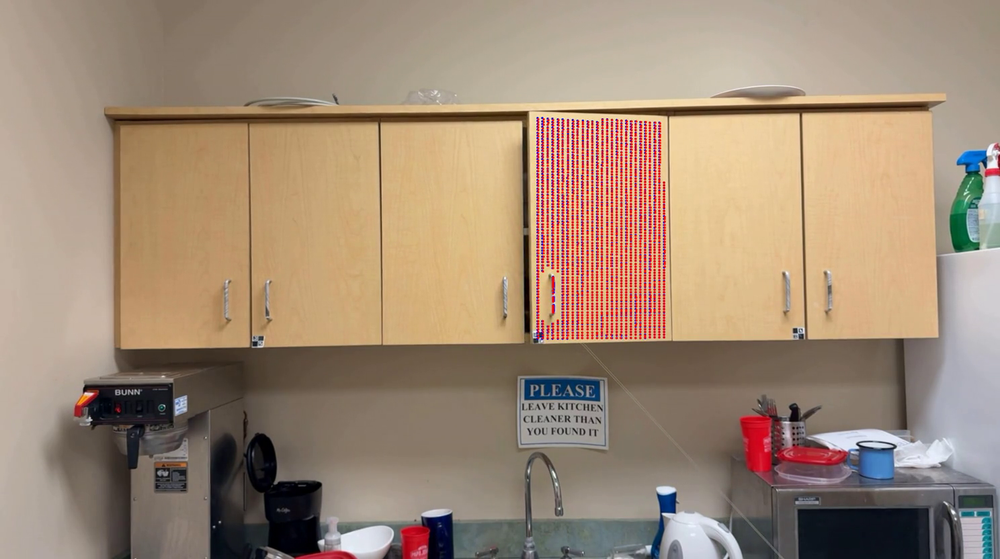} &
        \includegraphics[width=0.24\linewidth]{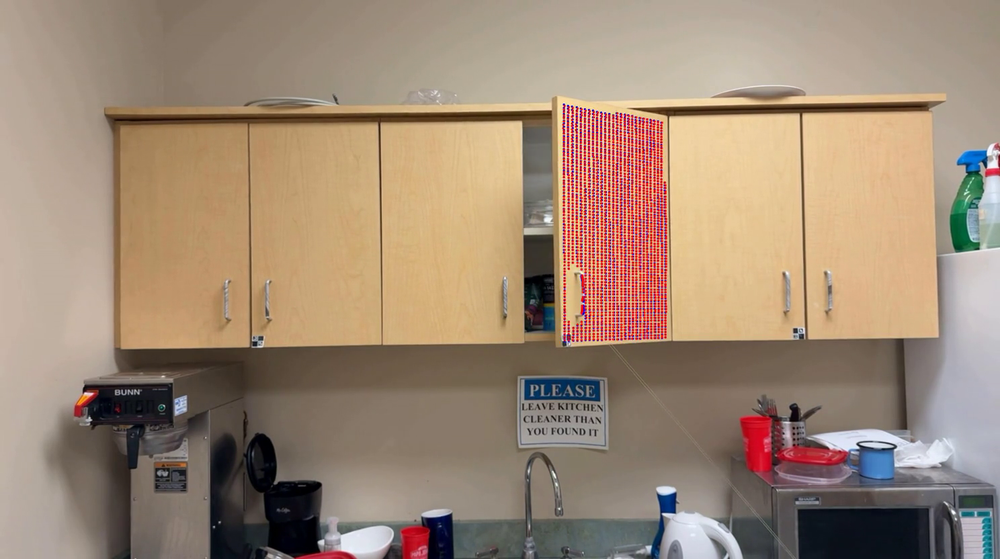} &
        \includegraphics[width=0.24\linewidth]{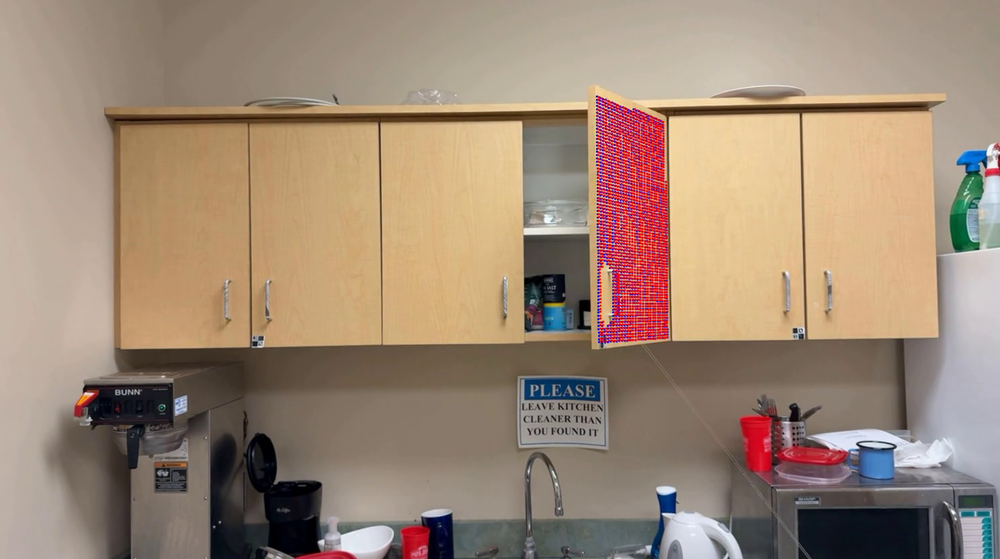} \\

    \end{tabular}
    }
    \vspace{-3mm}
    \captionof{figure}{
    \textbf{Qualitative Results on Simulated Motion Trajectories: } 
    We visualize our predicted articulation motion trajectories in \textcolor{blue}{blue}, and the ground truth trajectories in \textcolor{red}{red}.
    The close alignment between the two indicates the accuracy of our articulation estimation.
    }
    \label{fig:dense_traj}
\end{figure*}

\begin{table}
    \centering
    \setlength\tabcolsep{0.05em} %
    \vspace{2mm}
    \resizebox{0.48\textwidth}{!}{
    \begin{tabular}{cc}
          \animategraphics[autoplay,loop,width=0.48\linewidth, trim={0 0 0 0}, clip]{2}{figures/rebuttal/bathroom/}{001}{002}
          &\animategraphics[autoplay,loop,width=0.48\linewidth, trim={0 0 0 0}, clip]{2}{figures/rebuttal/office/}{001}{002}
    \end{tabular}
    }
    \vspace{-2mm}
    \captionof{figure}{\textbf{Beyond Kitchens:} DRAWER can generalize to different scenes such as offices and bedrooms. To view this figure as an \textbf{animation}, we recommend using Adobe Acrobat. 
    }
    \label{tab:more_scenes}
\end{table}

\begin{table*}[t]
    \centering
    \setlength\tabcolsep{0.05em} %
    \vspace{-2mm}\resizebox{\textwidth}{!}{
    \begin{tabular}{cccc}
          \includegraphics[width=0.25\textwidth]{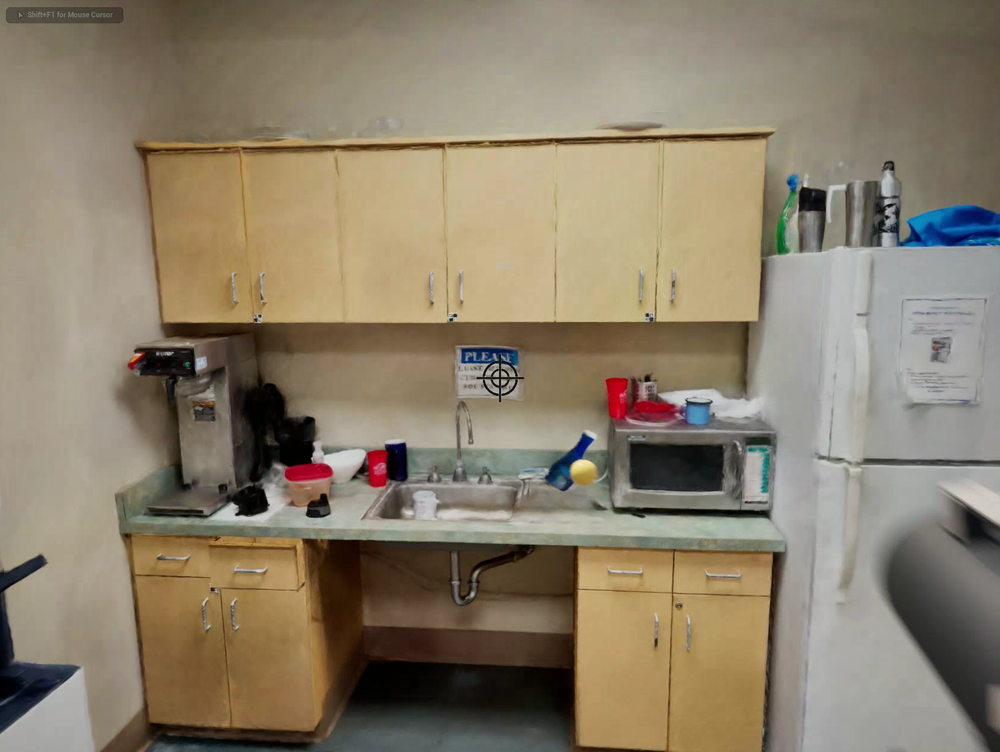} & 
          \includegraphics[width=0.25\textwidth]{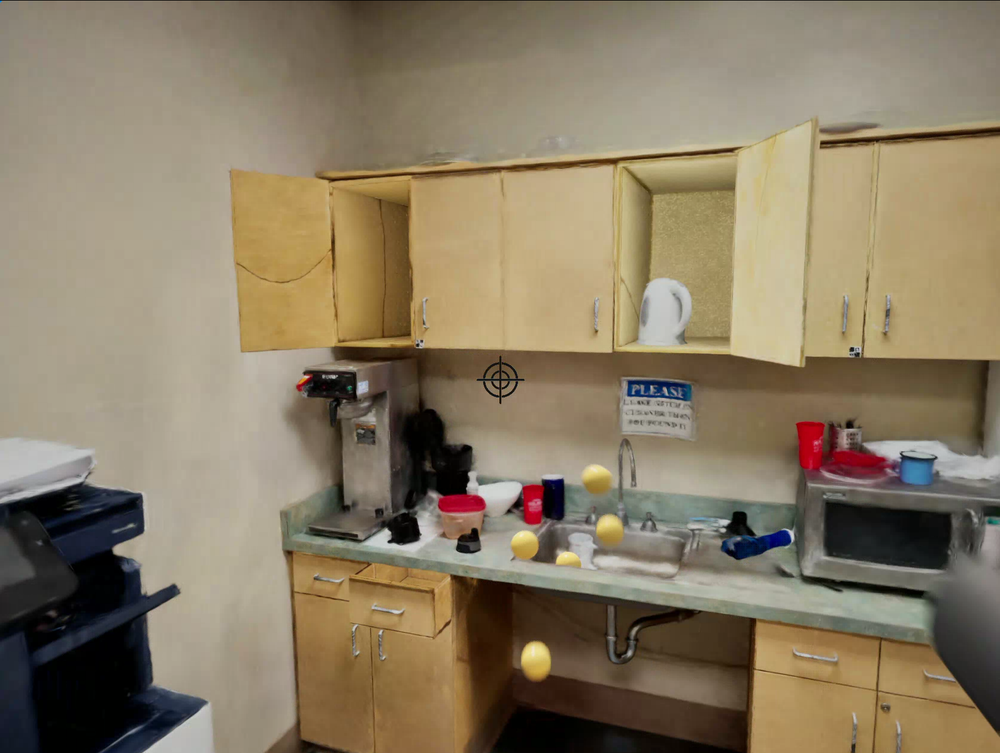} &
          \includegraphics[width=0.25\textwidth]{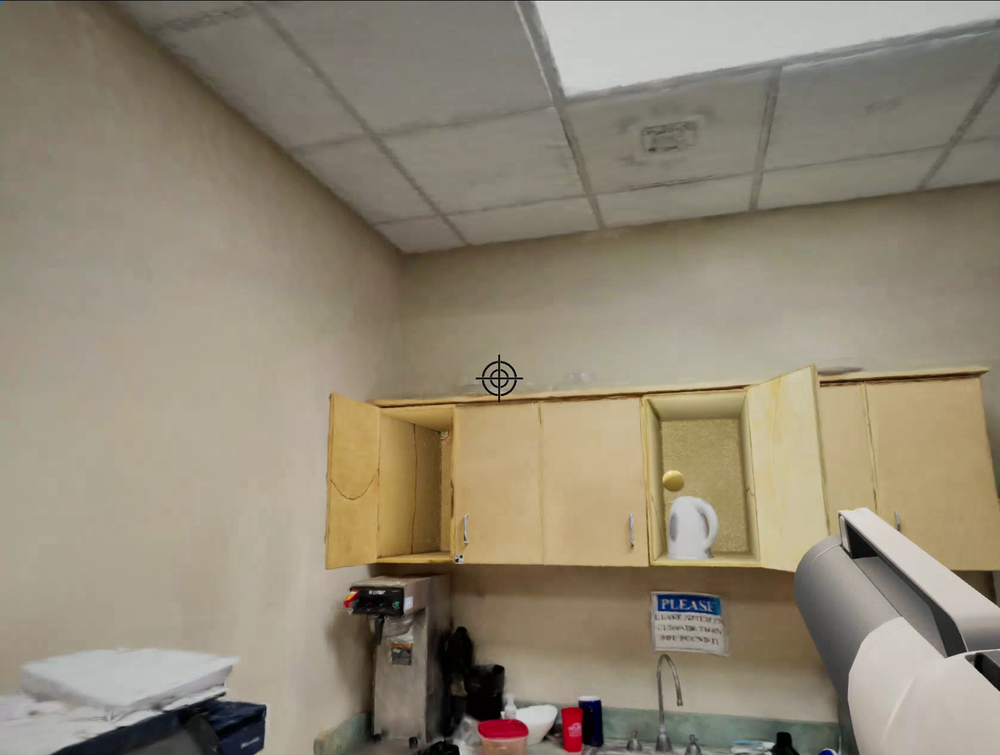} &
          \includegraphics[width=0.25\textwidth]{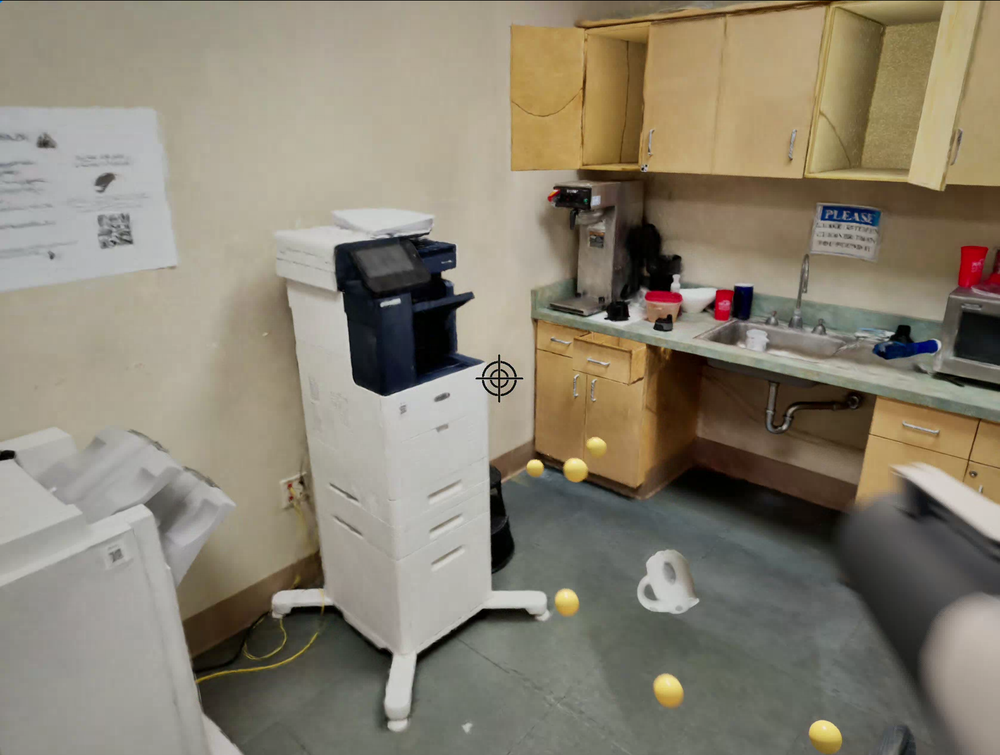} \\

    \end{tabular}
    }
    \vspace{-3mm}
    \captionof{figure}{{\bf \modelname ~in Unreal Engine.} {We demonstrate our interactive game in Unreal Engine with game features including shooting rigid objects like the blue bottle and white kettle segmented from the scene as well as opening cabinet and drawer doors.}}
    \label{tab:platform}
\end{table*}
\subsection{Articulating the Scene}
\label{sec:art_method}
Now that we have a dual scene representation of a \emph{static} scene, with high-quality visual appearance and detailed geometric structure, the next step is to estimate the underlying physical {properties}, such as articulation, and make the scene \emph{actionable} and \emph{interactable}. 

\paragraph{Scene decomposition:}
The first step is to identify potential interactable objects and segment them in 3D space. In this work, we focus on both articulated objects (\emph{e.g.}, drawers and cabinets) and rigid objects (\emph{e.g.}, cups and bottles). 
Here, we primarily discuss the processing of articulated objects, while details on rigid objects, which follow a similar pipeline, are provided in the supp. material.

Given a set of posed images, we first adopt {Grounded SAM} \cite{liu2023grounding, ren2024groundedsam, kirillov2023segment} to segment all objects of interest across all frames. 
Due to viewpoint variations and specular reflections, objects are not always fully visible, resulting in significant variation in mask quality. 
To filter out unreliable estimations and associate masks across frames, we {project} all masks onto the 3D mesh (obtained in Sec. \ref{sec:dual}) and fuse them using the Louvain algorithm \cite{traag2019louvain,zhang2021improved}. 
We discard masks whose IoU with their fusion results fall below a threshold.
Given the superior visual grounding capabilities of VLMs \cite{yang2023set}, we further employ GPT4o to assess mask quality and filter out unreliable ones.
We then exploit SAM \cite{kirillov2023segment} to re-segment objects using point prompts derived from the fusion results.
This process yields, for each object $i$, its high-quality 2D masks $\mathbf{M}_{i,j}$ in each image $j$, and its partial 3D geometry in either mesh form $\mathcal{M}^\text{obj}_i = (\mathbf{V}_i, \mathbf{F}_i)$ or SDF.

\paragraph{Physical reasoning:}
Once we identify all interactable objects, the next step is to estimate their physics-related attributes to effectively model and simulate their physical dynamics.
We adopt a two-pronged approach to estimate articulation types and axes of articulated objects.
The first prong leverages a specialized vision foundation model 3DOI \cite{qian2023understanding} to predict object hinges and affordances. We integrate its predictions with the underlying 3D to improve accuracy. 
Since 3DOI's performance varies by viewpoint, we use GPT4o for a more robust second estimation. When results differ, another VLM arbitrates the final decision. This  enhances overall accuracy.
For other physical parameters (\emph{e.g.}, mass, friction), following \cite{zhai2024physical,xia2024video2game}, one can either obtain estimates through VLM queries or set them manually.

\paragraph{Amodal shape estimation:} 
Having established object articulations, we now face the challenge of hidden regions.
When interacting with a reconstructed scene, previously invisible regions in the input video may become exposed. For instance, when cabinet doors open, their interior surfaces become visible. Without proper modeling, the geometry and appearance of these originally hidden regions would be under-constrained. This limitation prevents more sophisticated interactions, such as picking up a mug from the countertop and placing it into a drawer.

To address this issue, we first define a compositional 3D template for each object category $\mathcal{M}^\text{tmp} = (\mathbf{V}^\text{tmp}, \mathbf{F}^\text{tmp}) = \{(\mathbf{V}_i^\text{part}, \mathbf{F}_i^\text{part})\}_{i=1}^K$. Then, we exploit VLMs to refine the structure (\emph{e.g.}, adjusting the number of layers a cabinet has). Since the templates comprise well-defined shape primitives, we can easily edit the compositional structure or modify their shape to better match different observations.

For each object $i$, we consider three objectives: 
(i) the \emph{mask consistency term} measures the discrepancy between the rendered masks and the observed masks:  $\mathcal{L}_\text{mask} = \sum_j \lVert \mathbf{M}_{i,j} - \text{Rend}(\mathcal{M}^\text{tmp}_i) \rVert_2^2$; (ii) the \emph{shape consistency term} encourages the visible part of the template to match its corresponding partial 3D geometry in the scene. Since densely querying the learned SDF is computationally expensive, we instead adopt Chamfer Distance (CD): $\mathcal{L}_\text{shape} = \text{CD}(\mathcal{M}^\text{tmp, vis}_i, \mathcal{M}^\text{obj}_i)$; and (iii) the \emph{structure consistency term} encourages originally adjacent parts $\alpha, \beta$ to remain adjacent after optimization: $\mathcal{L}_\text{struc} = \sum_{(\alpha, \beta)} \sum_{(j, k)} \lVert \mathbf{V}^\alpha_{i,j} - \mathbf{V}^\beta_{i,k}\rVert_2^2$, where $(j, k)$ are the vertex pairs from $\alpha, \beta$ that are within a certain distance threshold.
We use PyTorch3D as the differentiable renderer. 
We optimize the poses and shape parameters (\emph{e.g.}, scale, width, length, etc) of all parts. 
We start with $\mathcal{L}_\text{shape}$ and $\mathcal{L}_\text{struc}$, and then turn on all objectives. 
To ensure objects do not collide with each other, we further adopt a {global} regularization term to penalize inter-penetration \cite{muller2021self}.

\paragraph{Texturing:} 
We exploit MatFuse \cite{vecchio2024matfuse}, a conditional diffusion model, to estimate the PBR materials of unobserved regions. We can either parameterize them as texture maps or distill them back to Gaussians in our dual representation.

\paragraph{Composing back to scene:} 
Directly editing and merging meshes is extremely difficult due to changes in topology. Fortunately, our dual representation, which builds on neural SDF and Gaussian splatting, is inherently compositional in 3D.  Furthermore, meshes and SDFs are largely interchangeable. 
Therefore, we can convert the completed articulated object back to the dual representation and use it to replace the original partial one.
\subsection{Downstream Applications with DRAWER}
\label{sec:interactive_env}

\paragraph{Gaming:} Content creation that reflects real world diversity with visual and physical fidelity, is a challenging problem for gaming applications. We demonstrate the utility of environments created with DRAWER for gaming applications using Unreal Engine (UE) \cite{unrealengine}. Specifically, we show how an agent dropped into the reconstructed environment imported into Unreal Engine can interact with the world to perform movement, shooting or opening of the various elements of the scene. Unreal engine offers support for rigid-body dynamics and articulation. Leveraging our dual representation—where Gaussians anchored to the SDF enable high-quality rendering, and SDF-derived mesh provides accurate collision geometry—we achieve alignment of rendering and physical models. The Luma Unreal Engine Plugin \cite{lumaaiunreal} allows real-time Gaussian rendering, while collision models use the SDF-extracted mesh. Articulation joints in UE are configured based on estimated types and axes, completing the interactive setup. As outlined in previous work~\cite{xia2024video2game}, we can then develop an interactive agent that can navigate and interact with various elements of the scene.

\paragraph{Real-to-sim-to-real transfer for robot learning:} 
Besides gaming applications, DRAWER holds value in data generation and model training for robotics. The environments created in DRAWER can be imported into a physics simulator such as Isaac Sim \cite{mittal2023orbit} with appropriate kinematics and dynamics. This enables the generation of physically realistic interaction data for tasks such as drawer opening/closing or object pick and place using motion planning or RL, without requiring tedious human effort. The generated data can be used to learn a policy that can be transferred to act in the real world directly from perception~\cite{torne2024reconciling, real2code, jiang2022ditto}. DRAWER generated environments offer greater visual and geometric fidelity than other environment creation methods, helping to bridge the simulation to reality gap. This circumvents much of the burden of real-world data collection on-robot.

%% file: sec/4_exp.tex
\begin{figure*}[t]
    \centering    \setlength\tabcolsep{0.05em} %
    \small    
    \vspace{-4mm}\resizebox{0.99\textwidth}{!}{
    \begin{tabular}{cccc}        
         \animategraphics[autoplay,loop,width=0.228\linewidth, trim={20 0 25 0}, clip]{20}{figures/video2/}{027}{133} &
        \includegraphics[width=0.25\textwidth]{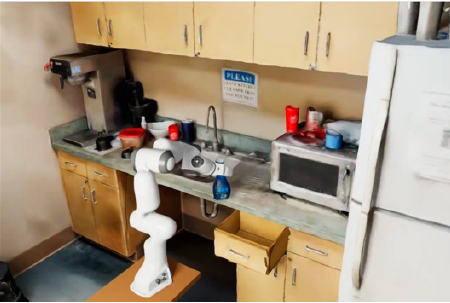} & 
        \includegraphics[width=0.25\textwidth]{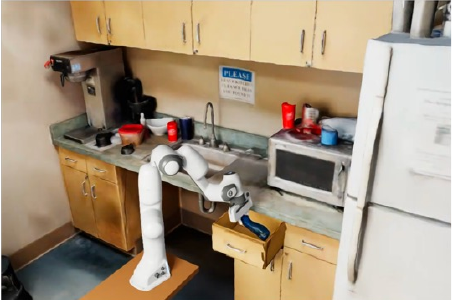} & 
        \includegraphics[width=0.25\textwidth]{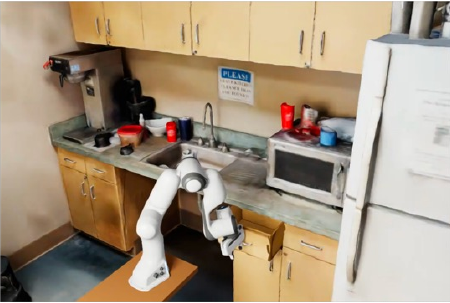} 
  
        \\
         Opening the drawer & Picking  & Placing & Closing the drawer
       
    \end{tabular}
    }
    \vspace{-3mm}\caption{\textbf{Scaling up robot learning data with our digital twin.} We demonstrate our interactive environment is capable of conducting robot learning tasks including opening/closing drawers and picking/placing segmented rigid objects in the scene. {To view this figure as a \textbf{video}, we recommend using Adobe Acrobat.}}
    \label{fig:sim-manipulation}
\end{figure*}

\section{Experiments}
\label{sec:exp}

\subsection{Setup}
\paragraph{Dataset:} 
\label{sec:dataset}
We manually capture videos in six different kitchens. 
The scenes are static, with no assumed interactions, which significantly reduces the capture cost. 
For evaluation, we annotate the type of articulation and any hinges present on all articulated objects within each scene. Additionally, we use fishing wire to open cabinet doors and drawers, obtaining GT video snippets of object articulations.
We further annotate key points on moving surfaces (\emph{e.g.}, cabinet doors, drawer fronts), fit homographies to these points, and extract dense 2D pixel trajectories. We use these trajectories to evaluate simulated articulated motions.

\paragraph{Metrics:} 
We adopt PSNR, SSIM, and LPIPS \cite{zhang2018perceptual} to assess the visual quality of digital twins. For articulation type estimation, we compute both precision and recall to explicitly account for potential perception errors (\emph{e.g.}, missed segmentation), which frequently occur in real-world scenarios. To quantify the accuracy of articulated motion simulation in digital twins, we track pixels along moving surfaces (\emph{e.g.}, cabinet doors) \cite{karaev2024cotracker3} and compute the Earth Mover's Distance (EMD) between simulated and ground truth trajectories. 
For revolute objects, we additionally report EA-Score \cite{zhao2021deep,qian2022understanding}, which measures both the angular and Euclidean distance between predicted and ground truth rotation axes.

\begin{table}[t]
\resizebox{0.48\textwidth}{!}{
    \begin{tabular}{lcccc}
       \toprule
       {{\textbf{Method}}} & {\textbf{Representation}} & {\textbf{PSNR}$\uparrow$} & {\textbf{SSIM}$\uparrow$}  & {\textbf{LPIPS}$\downarrow$}   \\
        \midrule
         Textured mesh 
         & Mesh & 19.71 & 0.785 & 0.323  \\
         \quad + Static GS 
         & Points+Mesh 
         & 25.87 & 0.889 & 0.276 \\
         \quad + Movable GS
         & Points+Mesh & 26.49 & 0.896 & 0.193 \\
         \quad + Straight-through est.
         & Points+Mesh & \textbf{27.80} & \textbf{0.912} & \textbf{0.159}  \\
        \bottomrule
    \end{tabular}
    }
    \caption{
    \textbf{Ablation Study on {\modelname} Model Design:} We observe improved rendering quality with the sequential addition of each designed component in {\modelname}.
    }
    \label{tab:ablation}
\end{table}

\begin{figure*}[t]
    \centering    \setlength\tabcolsep{0.05em} %
    \small    
    \vspace{-3mm}\resizebox{0.99\textwidth}{!}{
    \begin{tabular}{cccc}        
       
        \animategraphics[autoplay,loop,width=0.25\linewidth, trim={0 0 0 0}, clip]{20}{figures/video5/}{000}{062} &
        \includegraphics[width=0.25\textwidth]{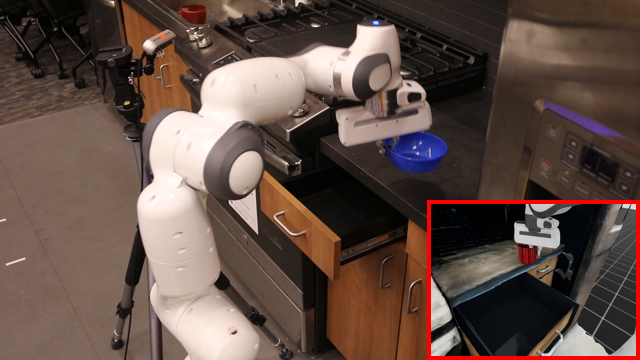} & 
        \includegraphics[width=0.25\textwidth]{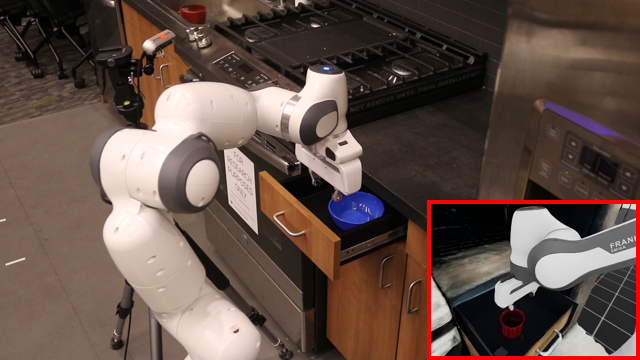} & 
        \includegraphics[width=0.25\textwidth]{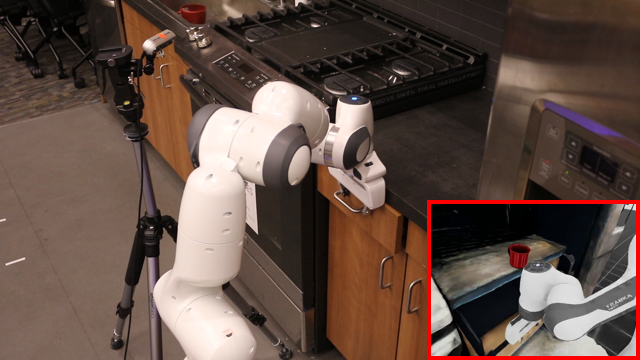} 
  
        \\
         Opening the drawer & Picking  & Placing & Closing the drawer
         \\

         \animategraphics[autoplay,loop,width=0.25\linewidth, trim={0 0 0 0}, clip]{20}{figures/deploy/revanimate/}{0000}{0026} &
        \includegraphics[width=0.25\textwidth]{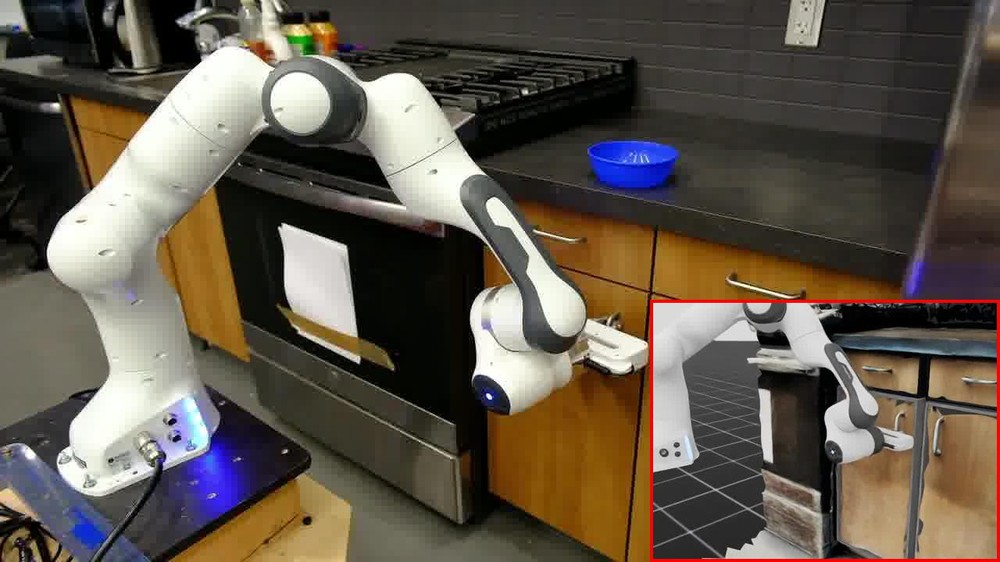} & 
        \includegraphics[width=0.25\textwidth]{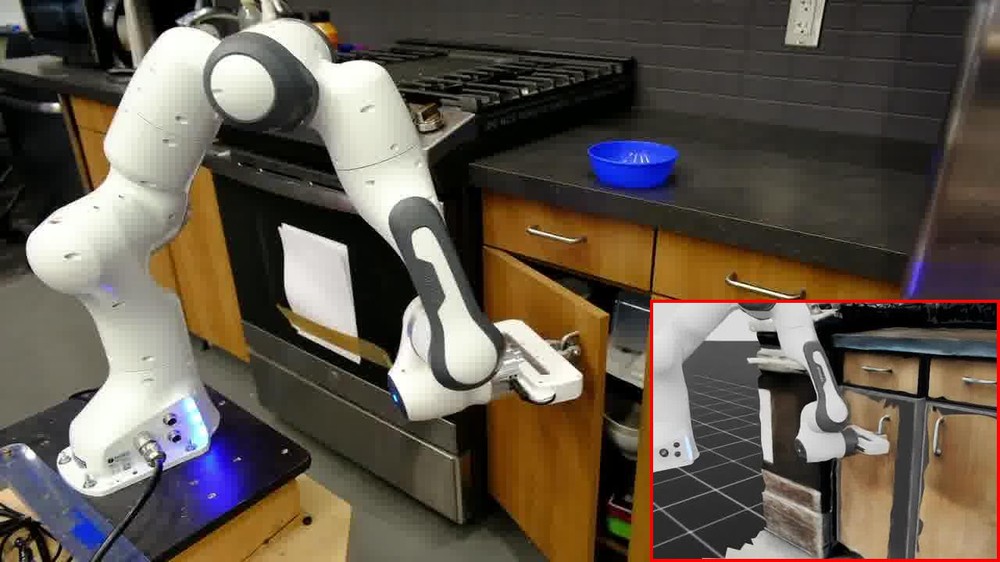} & 
        \includegraphics[width=0.25\textwidth]{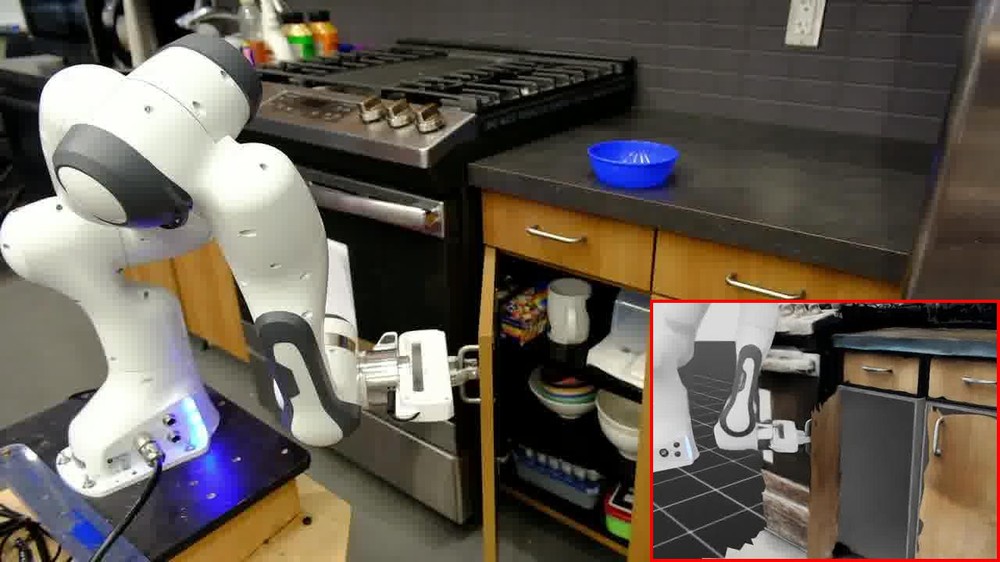} 
  
        \\
         Initial pose & Grasp handle  & Half open & Fully open
       
    \end{tabular}
    }
    \vspace{-3mm}\caption{\textbf{Real-to-Sim-to-Real.} {\modelname} allows us to learn  train robotic controllers using a real-to-sim-to-real loop. The inset images indicate the simulated data generation process.
    {To view this figure as a \textbf{video}, we recommend using Adobe Acrobat.}}
    \label{fig:deploy}
\end{figure*}

\subsection{Experimental Results}
\label{sec:exp_rs}
\paragraph{Novel view synthesis:}
Tab. \ref{tab:nvs} compares the visual realism and interactive capabilities of our reconstructed twins with those of prior works. We evaluate state-of-the-art approaches that utilize different representations, including Nerfacto \cite{tancik2023nerfstudio},  {BakedSDF} \cite{yariv2023bakedsdf}, {Video2Game} \cite{xia2024video2game}, 3DGS \cite{kerbl20233d}, and {2D GS} \cite{Huang2DGS2024}. 
Although 3DGS achieves the best rendering results, it compromises geometric fidelity, leading to reconstructed scenes that are neither realistic nor interactive. 
In contrast, our approach strikes the balance between the two. We significantly outperform neural rendering methods that support high-quality geometry, while also providing the most interactive functionalities, beyond simple pick-and-place.

\paragraph{Articulation estimation:} 
The performance of articulation estimation relies on both perception (\emph{i.e.}, identifying articulated) and reasoning (\emph{e.g.}, estimating articulation type).  
To validate the effectiveness of each module, we first assess the accuracy of our estimation \emph{given the masks of objects of interest}. 
We compare with {3DOI} \cite{qian2023understanding}, a foundation model for articulation prediction, in Tab. \ref{tab:quant-single-image-articulation} and Fig. \ref{fig:art-est}.
Since our approach leverages an ensemble of multi-modal models, it is more robust and achieves higher accuracy. By grounding predictions in the underlying 3D, we further improve the precision of estimated rotation axes.

We then evaluate the full pipeline, comparing it with recent methods for interactable 3D reconstruction: URDFormer \cite{chen2024urdformer} and Digital Cousin \cite{dai2024acdc}. 
As shown in Fig. \ref{fig:qual-digital-twin} and Tab. \ref{tab:quant-digital-twin-articulation}, our approach not only accurately recalls most articulated objects with high precision but also maintains exceptionally high visual and geometric fidelity. 
More results can be found in Fig. \ref{fig:more_results_kitchens}.

\paragraph{Articulated motion simulation:} 
Besides articulation type, the quality of simulated articulation motions is also crucial for creating high-quality digital twins. 
Since the reconstructions in \cite{chen2024urdformer,dai2024acdc} diverge significantly from the original scenes, they are not directly comparable. Instead, we compare our method with KlingAI \cite{klingai}, a SOTA conditional video diffusion model that supports motion control. 
We manually provide KlingAI point prompts to segment objects of interest and specify the desired articulated motions. 
Despite the privileged information, the synthesized motions from KlingAI are often infeasible (see Fig.~\ref{fig:qual-animate}). 
To further verify the fidelity of our simulated motions, we collect a set of ground truth articulated motion trajectories of opening drawers, cabinets, and refrigerators (see supp. materials for more details). 
We compute our motion trajectories by simulating the articulation in physics engines and we use CoTracker \cite{karaev2024cotracker3} to track the movement of KlingAI results. 
We use EMD to measure the distance among estimated and ground truth trajectories.
As shown in Fig \ref{fig:dense_traj}, our automatic approach is physically-grounded and outperforms KlingAI by an order of magnitude
 (EMD: $1.41\times10^{-5}$ \emph{v.s.} $17.7\times10^{-5}$).

\paragraph{Ablation study:}
We start with a {high-quality mesh} extracted from a neural SDF and sequentially add back other components. 
As shown in Tab. \ref{tab:ablation}, incorporating Gaussian splatting significantly improves overall performance. Additionally, allowing the Gaussians to move and using straight-through estimation further enhances the results. 

\paragraph{Beyond Kitchens:} While our primary focus is on kitchens -- due to their diversity of articulated and interactable objects -- DRAWER generalizes well to other scene types.
Fig.~\ref{tab:more_scenes} shows the performance of DRAWER on offices and bathrooms.

\subsection{Build Your Own Game}
\label{sec:video2game}

We have demonstrated our system’s effectiveness in rendering quality and articulation inference accuracy across various setups. Next, we construct an interactive game with first-person player control and real-time interaction.

\paragraph{Data preparation:} Our game assets are derived from self-captured kitchen videos (Sec.~\ref{sec:dataset}) and feature SDF-extracted mesh geometry for collision models, segmented objects for rigid-body dynamics, and articulated drawers with fully detailed interiors. Our dual-representation reconstruction enables high-quality Gaussian rendering in real time.

\paragraph{Interactive game features:} As shown in Fig. \ref{tab:platform}, our game, built upon Unreal Engine (UE) (Sec. \ref{sec:interactive_env}), supports high-quality rendering and diverse physical interactions at an interactive rate. Key features include:
\emph{Movement:} Players can navigate the room freely with realistic physics and collision models.
\emph{Shooting:} In a first-person view, players can shoot balls at segmented objects, with realistic motion simulated upon impact.
\emph{Opening:} Players can interact with articulated objects, such as drawers and cabinets, utilizing estimated articulations. Segmented items, like a kettle, can be removed from cabinets and dropped, with realistic dynamics and rendering enabled by our dual representation.

\begin{table}[t]
\resizebox{0.48\textwidth}{!}{
    \begin{tabular}{lccccc}
       \toprule
        Method & Total \# & Pred \# & Correct \#  & Precision (\%) & Recall (\%) \\
        \midrule
        URDFormer* \cite{chen2024urdformer} & \multirow{3}{*}{125} & 62 & 58 & 93.5 & 46.4\\
        Digital Cousin* \cite{dai2024acdc} &  & 160 & 75 & 46.9 & 60.0 \\
        Ours &  & 108 & 105 & \textbf{97.2} & \textbf{84.0} \\
        \bottomrule
    \end{tabular}
    }
    \vspace{-2mm}\caption{\textbf{Articulation Understanding %
    Comparison:}
    {Total\#, Pred\#, and Correct\# represent the total, predicted, and correctly predicted numbers of articulated objects.} {$^*$We conduct multiple predictions for each scene (based on different views) and select the best result.}
    }
    \label{tab:quant-digital-twin-articulation}
\end{table}

\begin{table}[t]
\resizebox{0.48\textwidth}{!}{
    \begin{tabular}{lccccc}
       \toprule
        Method & Total \# & Correct \#$\uparrow$ & Rev. \# & Correct Rev. \#$\uparrow$  & EA-Score $\uparrow$ \\
        \midrule
        3DOI & \multirow{2}{*}{80} & \textbf{78} & \multirow{2}{*}{59} & 57 & 0.861  \\
        Ours &  & \textbf{78} &  &  \textbf{58}  & \textbf{0.994} \\
        \bottomrule
    \end{tabular}
    }
    \vspace{-2mm}
    \caption{\textbf{Articulation Inference Comparison:}
    {
    Total \# and Correct \# represent the total number of articulated objects evaluated and the number correctly predicted, while Rev. \# denotes the count of correctly predicted revolute objects.    
    }
    }
    \label{tab:quant-single-image-articulation}
\end{table}

\subsection{Real-to-Sim-to-Real}
\label{sec:real2sim2real}

We conduct a proof of concept experiment with our articulated environment in a robotic real-to-sim-to-real setting. We reconstruct the scene with DRAWER, automatically generate simulation data via motion planning for policy learning, and transfer learned policies to the real world. A similar pipeline was shown in ~\citep{torne2024reconciling}, with manual articulations. 

\paragraph{Data generation:} 
To generate the data for policy learning, we first import the geometry and articulation reconstructed from the real scene via the DRAWER into Isaac Sim~\cite{mittal2023orbit}. As shown in Fig. \ref{fig:sim-manipulation}, we then initialize the robot's pose around each drawer and utilize standard motion planning~\cite{sundaralingam2023curobo} in combination with a object-centric grasp sampler~\cite{yuanm2t2} to generate motion data. This approach allows for generating physically realistic data for tasks such as opening the drawer by pulling the handle, picking and placing objects inside the drawer, and closing the drawer autonomously, without requiring considerable manual human effort. 

\paragraph{Policy learning:} Given this data, we then train a policy to open and close the drawer using behavior cloning on the collected data. Rendering the generated scenes in simulation as a point cloud, we deploy a commonly used policy learning architecture based on 3D Diffusion Policies~\cite{ze20243d}. Mirroring the deployment setting, the policy takes in a cropped point cloud and proprioceptive information and predicts the next end-effector pose of the robot that is then executed on the robot via inverse kinematics and position control.

\paragraph{Real-world deployment:} 
Finally, after training policies in simulation, we can directly transfer them to the real world on a Franka Emika Panda robot mounted on a mobile base. In this instantiation, we trained independent policies for each substage of the problem -- {drawer/cabinet opening, picking and placing, and closing.} A qualitative visualization of the learned behavior is shown in Fig~\ref{fig:deploy}. We refer readers to supp. material for more detailed visualizations.

%% file: sec/5_conclusion.tex
\section{Conclusion}
We present {\modelname}, a novel framework that automatically converts a single video into an interactive environment with articulated and rigid-body dynamics, requiring no prior articulation data. Our method integrates an SDF field and Gaussian splats into a dual scene representation, which is then decomposed and articulated to create a functional environment. We demonstrate {\modelname}'s superior performance in articulation understanding and rendering, as well as its utility in developing realistic interactive games and enabling real-to-sim-to-real transfer for robot learning. 
Looking ahead, {\modelname} could benefit from integration with more sophisticated relightable environment reconstruction.
\vspace{-7pt}
\paragraph{Acknowledgment:}
This research is partially supported by the Intel AI SRS gift, Meta research grant, Amazon, Army Research Lab, NVIDIA Academic Grant, the IBM IIDAI Grant, a gift from Ai2, NSF Awards \#2331878, \#2340254, \#2312102, \#2414227, and \#2404385, and DARPA TIAMAT Program \#HR00112490422.
The authors appreciate the NCSA for providing computing resources.

%% file: main.bbl
\begin{thebibliography}{115}
\providecommand{\natexlab}[1]{#1}
\providecommand{\url}[1]{\texttt{#1}}
\expandafter\ifx\csname urlstyle\endcsname\relax
  \providecommand{\doi}[1]{doi: #1}\else
  \providecommand{\doi}{doi: \begingroup \urlstyle{rm}\Url}\fi

\bibitem[Aliev et~al.(2020)Aliev, Sevastopolsky, Kolos, Ulyanov, and Lempitsky]{aliev2020neuralpointbasedgraphics}
Kara-Ali Aliev, Artem Sevastopolsky, Maria Kolos, Dmitry Ulyanov, and Victor Lempitsky.
\newblock Neural point-based graphics, 2020.

\bibitem[Amini et~al.(2022)Amini, Wang, Gilitschenski, Schwarting, Liu, Han, Karaman, and Rus]{amini2022vista}
Alexander Amini, Tsun-Hsuan Wang, Igor Gilitschenski, Wilko Schwarting, Zhijian Liu, Song Han, Sertac Karaman, and Daniela Rus.
\newblock Vista 2.0: An open, data-driven simulator for multimodal sensing and policy learning for autonomous vehicles.
\newblock In \emph{ICRA}, 2022.

\bibitem[Bae and Davison(2024)]{bae2024dsine}
Gwangbin Bae and Andrew~J. Davison.
\newblock Rethinking inductive biases for surface normal estimation.
\newblock In \emph{CVPR}, 2024.

\bibitem[Barron et~al.(2021)Barron, Mildenhall, Tancik, Hedman, Martin-Brualla, and Srinivasan]{barron2021mip}
Jonathan~T Barron, Ben Mildenhall, Matthew Tancik, Peter Hedman, Ricardo Martin-Brualla, and Pratul~P Srinivasan.
\newblock Mip-nerf: A multiscale representation for anti-aliasing neural radiance fields.
\newblock In \emph{ICCV}, 2021.

\bibitem[Barron et~al.(2022)Barron, Mildenhall, Verbin, Srinivasan, and Hedman]{barron2022mip}
Jonathan~T Barron, Ben Mildenhall, Dor Verbin, Pratul~P Srinivasan, and Peter Hedman.
\newblock Mip-nerf 360: Unbounded anti-aliased neural radiance fields.
\newblock In \emph{CVPR}, 2022.

\bibitem[Bengio et~al.(2013)Bengio, L{\'e}onard, and Courville]{bengio2013estimating}
Yoshua Bengio, Nicholas L{\'e}onard, and Aaron Courville.
\newblock Estimating or propagating gradients through stochastic neurons for conditional computation.
\newblock \emph{arXiv}, 2013.

\bibitem[Chen and Williams(1993)]{chen1993view}
Shenchang~Eric Chen and Lance Williams.
\newblock View interpolation for image synthesis.
\newblock In \emph{Proceedings of the 20th annual conference on Computer graphics and interactive techniques}, 1993.

\bibitem[Chen et~al.(2021)Chen, Rong, Duggal, Wang, Yan, Manivasagam, Xue, Yumer, and Urtasun]{chen2021geosim}
Yun Chen, Frieda Rong, Shivam Duggal, Shenlong Wang, Xinchen Yan, Sivabalan Manivasagam, Shangjie Xue, Ersin Yumer, and Raquel Urtasun.
\newblock Geosim: Realistic video simulation via geometry-aware composition for self-driving.
\newblock In \emph{CVPR}, 2021.

\bibitem[Chen et~al.(2023)Chen, Funkhouser, Hedman, and Tagliasacchi]{chen2022mobilenerf}
Zhiqin Chen, Thomas Funkhouser, Peter Hedman, and Andrea Tagliasacchi.
\newblock Mobilenerf: Exploiting the polygon rasterization pipeline for efficient neural field rendering on mobile architectures.
\newblock \emph{CVPR}, 2023.

\bibitem[Chen et~al.(2024)Chen, Walsman, Memmel, Mo, Fang, Vemuri, Wu, Fox, and Gupta]{chen2024urdformer}
Zoey Chen, Aaron Walsman, Marius Memmel, Kaichun Mo, Alex Fang, Karthikeya Vemuri, Alan Wu, Dieter Fox, and Abhishek Gupta.
\newblock Urdformer: A pipeline for constructing articulated simulation environments from real-world images.
\newblock \emph{arXiv}, 2024.

\bibitem[Dai et~al.(2024)Dai, Wong, Jiang, Wang, Gokmen, Zhang, Wu, and Fei-Fei]{dai2024acdc}
Tianyuan Dai, Josiah Wong, Yunfan Jiang, Chen Wang, Cem Gokmen, Ruohan Zhang, Jiajun Wu, and Li Fei-Fei.
\newblock Acdc: Automated creation of digital cousins for robust policy learning.
\newblock \emph{arXiv}, 2024.

\bibitem[Darmon et~al.(2022)Darmon, Bascle, Devaux, Monasse, and Aubry]{darmon2022improving}
Fran{\c{c}}ois Darmon, B{\'e}n{\'e}dicte Bascle, Jean-Cl{\'e}ment Devaux, Pascal Monasse, and Mathieu Aubry.
\newblock Improving neural implicit surfaces geometry with patch warping.
\newblock In \emph{Proceedings of the IEEE/CVF Conference on Computer Vision and Pattern Recognition}, pages 6260--6269, 2022.

\bibitem[Deitke et~al.(2022)Deitke, VanderBilt, Herrasti, Weihs, Ehsani, Salvador, Han, Kolve, Kembhavi, and Mottaghi]{ProcTHOR}
Matt Deitke, Eli VanderBilt, Alvaro Herrasti, Luca Weihs, Kiana Ehsani, Jordi Salvador, Winson Han, Eric Kolve, Aniruddha Kembhavi, and Roozbeh Mottaghi.
\newblock Procthor: Large-scale embodied ai using procedural generation.
\newblock \emph{NeurIPS}, 2022.

\bibitem[Dosovitskiy et~al.(2017)Dosovitskiy, Ros, Codevilla, Lopez, and Koltun]{dosovitskiy2017carla}
Alexey Dosovitskiy, German Ros, Felipe Codevilla, Antonio Lopez, and Vladlen Koltun.
\newblock Carla: An open urban driving simulator, 2017.

\bibitem[{Epic Games}()]{unrealengine}
{Epic Games}.
\newblock Unreal engine.

\bibitem[Games(2014)]{gta}
Rockstar Games.
\newblock Grand theft auto v, 2014.

\bibitem[Gao et~al.(2024)Gao, Yang, Zhang, Sun, Yuan, Fu, and Lai]{gao2024mesh}
Lin Gao, Jie Yang, Bo-Tao Zhang, Jia-Mu Sun, Yu-Jie Yuan, Hongbo Fu, and Yu-Kun Lai.
\newblock Mesh-based gaussian splatting for real-time large-scale deformation.
\newblock \emph{arXiv}, 2024.

\bibitem[Garcia et~al.(2024)Garcia, Zeid, Schmidt, de~Geus, Hermans, and Leibe]{garcia2024fine}
Gonzalo~Martin Garcia, Karim~Abou Zeid, Christian Schmidt, Daan de Geus, Alexander Hermans, and Bastian Leibe.
\newblock Fine-tuning image-conditional diffusion models is easier than you think.
\newblock \emph{arXiv}, 2024.

\bibitem[Gortler et~al.(1998)Gortler, He, Szeliski, et~al.]{gortler1998layered}
Jonathan Shade~Steven Gortler, Li-wei He, Richard Szeliski, et~al.
\newblock Layered depth images.
\newblock In \emph{SIGGRAPH}, pages 231--242, 1998.

\bibitem[Guo et~al.(2023)Guo, Yang, Rao, Liang, Wang, Qiao, Agrawala, Lin, and Dai]{guo2023animatediff}
Yuwei Guo, Ceyuan Yang, Anyi Rao, Zhengyang Liang, Yaohui Wang, Yu Qiao, Maneesh Agrawala, Dahua Lin, and Bo Dai.
\newblock Animatediff: Animate your personalized text-to-image diffusion models without specific tuning, 2023.

\bibitem[Guédon and Lepetit(2023)]{guédon2023sugarsurfacealignedgaussiansplatting}
Antoine Guédon and Vincent Lepetit.
\newblock Sugar: Surface-aligned gaussian splatting for efficient 3d mesh reconstruction and high-quality mesh rendering, 2023.

\bibitem[Hao et~al.(2018)Hao, Huang, and Belongie]{hao2018controllable}
Zekun Hao, Xun Huang, and Serge Belongie.
\newblock Controllable video generation with sparse trajectories.
\newblock In \emph{CVPR}, 2018.

\bibitem[Hayden et~al.(2020)Hayden, Pacheco, and Fisher]{hayden2020nonparametric}
David~S Hayden, Jason Pacheco, and John~W Fisher.
\newblock Nonparametric object and parts modeling with lie group dynamics.
\newblock In \emph{CVPR}, 2020.

\bibitem[Heigl et~al.(1999)Heigl, Koch, Pollefeys, Denzler, and Van~Gool]{heigl1999plenoptic}
Benno Heigl, Reinhard Koch, Marc Pollefeys, Joachim Denzler, and Luc Van~Gool.
\newblock Plenoptic modeling and rendering from image sequences taken by a hand-held camera.
\newblock In \emph{DAGM-Symposium}, 1999.

\bibitem[Hsu et~al.(2023)Hsu, Jiang, and Zhu]{hsu2023ditto}
Cheng-Chun Hsu, Zhenyu Jiang, and Yuke Zhu.
\newblock Ditto in the house: Building articulation models of indoor scenes through interactive perception.
\newblock In \emph{ICRA}, 2023.

\bibitem[Hsu et~al.(2024)Hsu, Lin, Zhai, Xia, and Wang]{hsu2024autovfxphysicallyrealisticvideo}
Hao-Yu Hsu, Zhi-Hao Lin, Albert Zhai, Hongchi Xia, and Shenlong Wang.
\newblock Autovfx: Physically realistic video editing from natural language instructions, 2024.

\bibitem[Huang et~al.(2024)Huang, Yu, Chen, Geiger, and Gao]{Huang2DGS2024}
Binbin Huang, Zehao Yu, Anpei Chen, Andreas Geiger, and Shenghua Gao.
\newblock 2d gaussian splatting for geometrically accurate radiance fields.
\newblock In \emph{SIGGRAPH}. Association for Computing Machinery, 2024.

\bibitem[Huang et~al.(2021)Huang, Wang, Birdal, Sung, Arrigoni, Hu, and Guibas]{huang2021multibodysync}
Jiahui Huang, He Wang, Tolga Birdal, Minhyuk Sung, Federica Arrigoni, Shi-Min Hu, and Leonidas~J Guibas.
\newblock Multibodysync: Multi-body segmentation and motion estimation via 3d scan synchronization.
\newblock In \emph{CVPR}, 2021.

\bibitem[Iliash et~al.(2024)Iliash, Jiang, Zhang, Savva, and Chang]{iliash2024s2o}
Denys Iliash, Hanxiao Jiang, Yiming Zhang, Manolis Savva, and Angel~X Chang.
\newblock S2o: Static to openable enhancement for articulated 3d objects.
\newblock \emph{arXiv}, 2024.

\bibitem[Jiang et~al.(2022)Jiang, Hsu, and Zhu]{jiang2022ditto}
Zhenyu Jiang, Cheng-Chun Hsu, and Yuke Zhu.
\newblock Ditto: Building digital twins of articulated objects from interaction.
\newblock In \emph{CVPR}, 2022.

\bibitem[Karaev et~al.(2024)Karaev, Makarov, Wang, Neverova, Vedaldi, and Rupprecht]{karaev2024cotracker3}
Nikita Karaev, Iurii Makarov, Jianyuan Wang, Natalia Neverova, Andrea Vedaldi, and Christian Rupprecht.
\newblock Cotracker3: Simpler and better point tracking by pseudo-labelling real videos.
\newblock \emph{arXiv}, 2024.

\bibitem[Kerbl et~al.(2023)Kerbl, Kopanas, Leimk{\"u}hler, and Drettakis]{kerbl20233d}
Bernhard Kerbl, Georgios Kopanas, Thomas Leimk{\"u}hler, and George Drettakis.
\newblock 3d gaussian splatting for real-time radiance field rendering.
\newblock \emph{TOG}, 2023.

\bibitem[Kerr et~al.(2024)Kerr, Kim, Wu, Yi, Wang, Goldberg, and Kanazawa]{kerr2024robot}
Justin Kerr, Chung~Min Kim, Mingxuan Wu, Brent Yi, Qianqian Wang, Ken Goldberg, and Angjoo Kanazawa.
\newblock Robot see robot do: Imitating articulated object manipulation with monocular 4d reconstruction.
\newblock \emph{arXiv preprint arXiv:2409.18121}, 2024.

\bibitem[Kirillov et~al.(2023)Kirillov, Mintun, Ravi, Mao, Rolland, Gustafson, Xiao, Whitehead, Berg, Lo, et~al.]{kirillov2023segment}
Alexander Kirillov, Eric Mintun, Nikhila Ravi, Hanzi Mao, Chloe Rolland, Laura Gustafson, Tete Xiao, Spencer Whitehead, Alexander~C Berg, Wan-Yen Lo, et~al.
\newblock Segment anything.
\newblock In \emph{ICCV}, 2023.

\bibitem[KlingAI(2024)]{klingai}
KlingAI, 2024.

\bibitem[Le et~al.(2024)Le, Xie, Liang, Wang, Yang, Ma, Vedder, Krishna, Jayaraman, and Eaton]{le2024articulate}
Long Le, Jason Xie, William Liang, Hung-Ju Wang, Yue Yang, Yecheng~Jason Ma, Kyle Vedder, Arjun Krishna, Dinesh Jayaraman, and Eric Eaton.
\newblock Articulate-anything: Automatic modeling of articulated objects via a vision-language foundation model.
\newblock \emph{arXiv}, 2024.

\bibitem[Lei et~al.(2023)Lei, Deng, Shen, Guibas, and Daniilidis]{lei2023nap}
Jiahui Lei, Congyue Deng, William~B Shen, Leonidas~J Guibas, and Kostas Daniilidis.
\newblock Nap: Neural 3d articulated object prior.
\newblock \emph{NeurIPS}, 2023.

\bibitem[Levoy and Hanrahan(1996)]{levoy1996light}
Marc Levoy and Pat Hanrahan.
\newblock Light field rendering.
\newblock In \emph{Proceedings of the 23rd annual conference on Computer graphics and interactive techniques}, 1996.

\bibitem[Li et~al.(2022)Li, Lin, Forsyth, Huang, and Wang]{li2022climatenerf}
Yuan Li, Zhi-Hao Lin, David Forsyth, Jia-Bin Huang, and Shenlong Wang.
\newblock Climatenerf: Physically-based neural rendering for extreme climate synthesis.
\newblock \emph{arXiv}, 2022.

\bibitem[Li et~al.(2023)Li, M{\"u}ller, Evans, Taylor, Unberath, Liu, and Lin]{li2023neuralangelo}
Zhaoshuo Li, Thomas M{\"u}ller, Alex Evans, Russell~H Taylor, Mathias Unberath, Ming-Yu Liu, and Chen-Hsuan Lin.
\newblock Neuralangelo: High-fidelity neural surface reconstruction.
\newblock In \emph{Proceedings of the IEEE/CVF Conference on Computer Vision and Pattern Recognition}, pages 8456--8465, 2023.

\bibitem[Li et~al.(2024)Li, Tucker, Snavely, and Holynski]{li2024generative}
Zhengqi Li, Richard Tucker, Noah Snavely, and Aleksander Holynski.
\newblock Generative image dynamics.
\newblock In \emph{CVPR}, 2024.

\bibitem[Lin et~al.(2022)Lin, Ma, Hsu, Wang, and Wang]{lin2022neurmips}
Zhi-Hao Lin, Wei-Chiu Ma, Hao-Yu Hsu, Yu-Chiang~Frank Wang, and Shenlong Wang.
\newblock Neurmips: Neural mixture of planar experts for view synthesis.
\newblock In \emph{CVPR}, 2022.

\bibitem[Lin et~al.(2023)Lin, Liu, Chen, Forsyth, Huang, Bhattad, and Wang]{lin2023urbanir}
Zhi-Hao Lin, Bohan Liu, Yi-Ting Chen, David Forsyth, Jia-Bin Huang, Anand Bhattad, and Shenlong Wang.
\newblock Urbanir: Large-scale urban scene inverse rendering from a single video, 2023.

\bibitem[Liu et~al.(2024{\natexlab{a}})Liu, Iliash, Chang, Savva, and Mahdavi-Amiri]{liu2024singapo}
Jiayi Liu, Denys Iliash, Angel~X Chang, Manolis Savva, and Ali Mahdavi-Amiri.
\newblock Singapo: Single image controlled generation of articulated parts in object.
\newblock \emph{arXiv}, 2024{\natexlab{a}}.

\bibitem[Liu et~al.(2023{\natexlab{a}})Liu, Chen, Yang, Wang, Manivasagam, and Urtasun]{liu2023real}
Jeffrey~Yunfan Liu, Yun Chen, Ze Yang, Jingkang Wang, Sivabalan Manivasagam, and Raquel Urtasun.
\newblock Real-time neural rasterization for large scenes.
\newblock In \emph{Proceedings of the IEEE/CVF International Conference on Computer Vision}, pages 8416--8427, 2023{\natexlab{a}}.

\bibitem[Liu et~al.(2023{\natexlab{b}})Liu, Gupta, and Wang]{liu2023building}
Shaowei Liu, Saurabh Gupta, and Shenlong Wang.
\newblock Building rearticulable models for arbitrary 3d objects from 4d point clouds.
\newblock In \emph{Proceedings of the IEEE/CVF Conference on Computer Vision and Pattern Recognition}, pages 21138--21147, 2023{\natexlab{b}}.

\bibitem[Liu et~al.(2023{\natexlab{c}})Liu, Zeng, Ren, Li, Zhang, Yang, Li, Yang, Su, Zhu, et~al.]{liu2023grounding}
Shilong Liu, Zhaoyang Zeng, Tianhe Ren, Feng Li, Hao Zhang, Jie Yang, Chunyuan Li, Jianwei Yang, Hang Su, Jun Zhu, et~al.
\newblock Grounding dino: Marrying dino with grounded pre-training for open-set object detection.
\newblock \emph{arXiv}, 2023{\natexlab{c}}.

\bibitem[Liu et~al.(2024{\natexlab{b}})Liu, Ren, Gupta, and Wang]{liu2024physgenrigidbodyphysicsgroundedimagetovideo}
Shaowei Liu, Zhongzheng Ren, Saurabh Gupta, and Shenlong Wang.
\newblock Physgen: Rigid-body physics-grounded image-to-video generation, 2024{\natexlab{b}}.

\bibitem[Liu et~al.(2025)Liu, Jia, Lu, Ni, Zhu, and Huang]{liu2025building}
Yu Liu, Baoxiong Jia, Ruijie Lu, Junfeng Ni, Song-Chun Zhu, and Siyuan Huang.
\newblock Building interactable replicas of complex articulated objects via gaussian splatting.
\newblock \emph{arXiv preprint arXiv:2502.19459}, 2025.

\bibitem[Liu et~al.(2023{\natexlab{d}})Liu, Zhou, He, Marcucci, Fei-Fei, Wu, and Li]{liu2023modelbasedcontrolsparseneural}
Ziang Liu, Genggeng Zhou, Jeff He, Tobia Marcucci, Li Fei-Fei, Jiajun Wu, and Yunzhu Li.
\newblock Model-based control with sparse neural dynamics, 2023{\natexlab{d}}.

\bibitem[Lu et~al.(2023)Lu, Xu, Chen, Li, Lin, and Jiang]{lu2023urban}
Fan Lu, Yan Xu, Guang Chen, Hongsheng Li, Kwan-Yee Lin, and Changjun Jiang.
\newblock Urban radiance field representation with deformable neural mesh primitives, 2023.

\bibitem[Luiten et~al.(2024)Luiten, Kopanas, Leibe, and Ramanan]{luiten2023dynamic}
Jonathon Luiten, Georgios Kopanas, Bastian Leibe, and Deva Ramanan.
\newblock Dynamic 3d gaussians: Tracking by persistent dynamic view synthesis.
\newblock In \emph{3DV}, 2024.

\bibitem[{Luma AI}()]{lumaaiunreal}
{Luma AI}.
\newblock Luma unreal engine plugin.

\bibitem[Mandi et~al.(2024{\natexlab{a}})Mandi, Weng, Bauer, and Song]{mandi2024real2code}
Zhao Mandi, Yijia Weng, Dominik Bauer, and Shuran Song.
\newblock Real2code: Reconstruct articulated objects via code generation.
\newblock \emph{arXiv}, 2024{\natexlab{a}}.

\bibitem[Mandi et~al.(2024{\natexlab{b}})Mandi, Weng, Bauer, and Song]{real2code}
Zhao Mandi, Yijia Weng, Dominik Bauer, and Shuran Song.
\newblock Real2code: Reconstruct articulated objects via code generation.
\newblock \emph{CoRR}, abs/2406.08474, 2024{\natexlab{b}}.

\bibitem[Manivasagam et~al.(2020)Manivasagam, Wang, Wong, Zeng, Sazanovich, Tan, Yang, Ma, and Urtasun]{manivasagam2020lidarsim}
Sivabalan Manivasagam, Shenlong Wang, Kelvin Wong, Wenyuan Zeng, Mikita Sazanovich, Shuhan Tan, Bin Yang, Wei-Chiu Ma, and Raquel Urtasun.
\newblock Lidarsim: Realistic lidar simulation by leveraging the real world.
\newblock In \emph{CVPR}, 2020.

\bibitem[Max(1995)]{max1995optical}
Nelson Max.
\newblock Optical models for direct volume rendering.
\newblock \emph{TOG}, 1995.

\bibitem[Mildenhall et~al.(2020)Mildenhall, Srinivasan, Tancik, Barron, Ramamoorthi, and Ng]{mildenhall2020nerf}
Ben Mildenhall, Pratul~P. Srinivasan, Matthew Tancik, Jonathan~T. Barron, Ravi Ramamoorthi, and Ren Ng.
\newblock Nerf: Representing scenes as neural radiance fields for view synthesis.
\newblock In \emph{ECCV}, 2020.

\bibitem[Mildenhall et~al.(2021)Mildenhall, Srinivasan, Tancik, Barron, Ramamoorthi, and Ng]{mildenhall2021nerf}
Ben Mildenhall, Pratul~P Srinivasan, Matthew Tancik, Jonathan~T Barron, Ravi Ramamoorthi, and Ren Ng.
\newblock Nerf: Representing scenes as neural radiance fields for view synthesis.
\newblock \emph{ACM Communications}, 2021.

\bibitem[Mildenhall et~al.(2022)Mildenhall, Hedman, Martin-Brualla, Srinivasan, and Barron]{mildenhall2022nerf}
Ben Mildenhall, Peter Hedman, Ricardo Martin-Brualla, Pratul~P Srinivasan, and Jonathan~T Barron.
\newblock Nerf in the dark: High dynamic range view synthesis from noisy raw images.
\newblock In \emph{Proceedings of the IEEE/CVF Conference on Computer Vision and Pattern Recognition}, pages 16190--16199, 2022.

\bibitem[Mittal et~al.(2023)Mittal, Yu, Yu, Liu, Rudin, Hoeller, Yuan, Singh, Guo, Mazhar, Mandlekar, Babich, State, Hutter, and Garg]{mittal2023orbit}
Mayank Mittal, Calvin Yu, Qinxi Yu, Jingzhou Liu, Nikita Rudin, David Hoeller, Jia~Lin Yuan, Ritvik Singh, Yunrong Guo, Hammad Mazhar, Ajay Mandlekar, Buck Babich, Gavriel State, Marco Hutter, and Animesh Garg.
\newblock Orbit: A unified simulation framework for interactive robot learning environments.
\newblock \emph{IEEE Robotics and Automation Letters}, 2023.

\bibitem[Muller et~al.(2021)Muller, Osman, Tang, Huang, and Black]{muller2021self}
Lea Muller, Ahmed~AA Osman, Siyu Tang, Chun-Hao~P Huang, and Michael~J Black.
\newblock On self-contact and human pose.
\newblock In \emph{CVPR}, 2021.

\bibitem[Müller et~al.(2022)Müller, Evans, Schied, and Keller]{M_ller_2022}
Thomas Müller, Alex Evans, Christoph Schied, and Alexander Keller.
\newblock Instant neural graphics primitives with a multiresolution hash encoding.
\newblock \emph{ACM Transactions on Graphics}, 41\penalty0 (4):\penalty0 1–15, 2022.

\bibitem[Noguchi et~al.(2022)Noguchi, Iqbal, Tremblay, Harada, and Gallo]{noguchi2022watch}
Atsuhiro Noguchi, Umar Iqbal, Jonathan Tremblay, Tatsuya Harada, and Orazio Gallo.
\newblock Watch it move: Unsupervised discovery of 3d joints for re-posing of articulated objects.
\newblock In \emph{CVPR}, 2022.

\bibitem[Oechsle et~al.(2021)Oechsle, Peng, and Geiger]{oechsle2021unisurf}
Michael Oechsle, Songyou Peng, and Andreas Geiger.
\newblock Unisurf: Unifying neural implicit surfaces and radiance fields for multi-view reconstruction.
\newblock In \emph{Proceedings of the IEEE/CVF International Conference on Computer Vision}, pages 5589--5599, 2021.

\bibitem[Paudel et~al.(2024)Paudel, Khanal, Chhatkuli, Paudel, and Tandukar]{paudel2024ihuman}
Pramish Paudel, Anubhav Khanal, Ajad Chhatkuli, Danda~Pani Paudel, and Jyoti Tandukar.
\newblock ihuman: Instant animatable digital humans from monocular videos.
\newblock \emph{arXiv}, 2024.

\bibitem[Pun et~al.(2023)Pun, Sun, Wang, Chen, Yang, Manivasagam, Ma, and Urtasun]{pun2023lightsimneurallightingsimulation}
Ava Pun, Gary Sun, Jingkang Wang, Yun Chen, Ze Yang, Sivabalan Manivasagam, Wei-Chiu Ma, and Raquel Urtasun.
\newblock Lightsim: Neural lighting simulation for urban scenes, 2023.

\bibitem[Pun et~al.(2024)Pun, Sun, Wang, Chen, Yang, Manivasagam, Ma, and Urtasun]{pun2024neural}
Ava Pun, Gary Sun, Jingkang Wang, Yun Chen, Ze Yang, Sivabalan Manivasagam, Wei-Chiu Ma, and Raquel Urtasun.
\newblock Neural lighting simulation for urban scenes.
\newblock \emph{NeurIPS}, 2024.

\bibitem[Qian and Fouhey(2023)]{qian2023understanding}
Shengyi Qian and David~F Fouhey.
\newblock Understanding 3d object interaction from a single image.
\newblock In \emph{ICCV}, 2023.

\bibitem[Qian et~al.(2022)Qian, Jin, Rockwell, Chen, and Fouhey]{qian2022understanding}
Shengyi Qian, Linyi Jin, Chris Rockwell, Siyi Chen, and David~F Fouhey.
\newblock Understanding 3d object articulation in internet videos.
\newblock In \emph{CVPR}, 2022.

\bibitem[Qian et~al.(2024)Qian, Kirschstein, Schoneveld, Davoli, Giebenhain, and Nießner]{qian2024gaussianavatars}
Shenhan Qian, Tobias Kirschstein, Liam Schoneveld, Davide Davoli, Simon Giebenhain, and Matthias Nießner.
\newblock Gaussianavatars: Photorealistic head avatars with rigged 3d gaussians.
\newblock \emph{arXiv}, 2024.

\bibitem[Raistrick et~al.(2024)Raistrick, Mei, Kayan, Yan, Zuo, Han, Wen, Parakh, Alexandropoulos, Lipson, Ma, and Deng]{raistrick2024infinigenindoorsphotorealisticindoor}
Alexander Raistrick, Lingjie Mei, Karhan Kayan, David Yan, Yiming Zuo, Beining Han, Hongyu Wen, Meenal Parakh, Stamatis Alexandropoulos, Lahav Lipson, Zeyu Ma, and Jia Deng.
\newblock Infinigen indoors: Photorealistic indoor scenes using procedural generation, 2024.

\bibitem[Ren et~al.(2024)Ren, Liu, Zeng, Lin, Li, Cao, Chen, Huang, Chen, Yan, Zeng, Zhang, Li, Yang, Li, Jiang, and Zhang]{ren2024groundedsam}
Tianhe Ren, Shilong Liu, Ailing Zeng, Jing Lin, Kunchang Li, He Cao, Jiayu Chen, Xinyu Huang, Yukang Chen, Feng Yan, Zhaoyang Zeng, Hao Zhang, Feng Li, Jie Yang, Hongyang Li, Qing Jiang, and Lei Zhang.
\newblock Grounded sam: Assembling open-world models for diverse visual tasks, 2024.

\bibitem[Son et~al.(2022)Son, Qiao, Sewall, and Lin]{son2022differentiable}
Sanghyun Son, Yi-Ling Qiao, Jason Sewall, and Ming~C Lin.
\newblock Differentiable hybrid traffic simulation.
\newblock \emph{TOG}, 2022.

\bibitem[Sundaralingam et~al.(2023)Sundaralingam, Hari, Fishman, Garrett, Van~Wyk, Blukis, Millane, Oleynikova, Handa, Ramos, et~al.]{sundaralingam2023curobo}
Balakumar Sundaralingam, Siva Kumar~Sastry Hari, Adam Fishman, Caelan Garrett, Karl Van~Wyk, Valts Blukis, Alexander Millane, Helen Oleynikova, Ankur Handa, Fabio Ramos, et~al.
\newblock Curobo: Parallelized collision-free robot motion generation.
\newblock In \emph{2023 IEEE International Conference on Robotics and Automation (ICRA)}, pages 8112--8119. IEEE, 2023.

\bibitem[Szeliski and Golland(1998)]{szeliski1998stereo}
Richard Szeliski and Polina Golland.
\newblock Stereo matching with transparency and matting.
\newblock In \emph{Sixth International Conference on Computer Vision}, 1998.

\bibitem[Tancik et~al.(2023)Tancik, Weber, Ng, Li, Yi, Kerr, Wang, Kristoffersen, Austin, Salahi, et~al.]{tancik2023nerfstudio}
Matthew Tancik, Ethan Weber, Evonne Ng, Ruilong Li, Brent Yi, Justin Kerr, Terrance Wang, Alexander Kristoffersen, Jake Austin, Kamyar Salahi, et~al.
\newblock Nerfstudio: A modular framework for neural radiance field development.
\newblock \emph{in arXiv}, 2023.

\bibitem[Tang et~al.(2023)Tang, Zhou, Chen, Hu, Ding, Wang, and Zeng]{tang2022nerf2mesh}
Jiaxiang Tang, Hang Zhou, Xiaokang Chen, Tianshu Hu, Errui Ding, Jingdong Wang, and Gang Zeng.
\newblock Delicate textured mesh recovery from nerf via adaptive surface refinement.
\newblock \emph{arXiv preprint arXiv:2303.02091}, 2023.

\bibitem[Todorov et~al.(2012)Todorov, Erez, and Tassa]{todorov2012mujoco}
Emanuel Todorov, Tom Erez, and Yuval Tassa.
\newblock Mujoco: A physics engine for model-based control.
\newblock In \emph{IROS}, 2012.

\bibitem[Torne et~al.(2024)Torne, Simeonov, Li, Chan, Chen, Gupta, and Agrawal]{torne2024reconciling}
Marcel Torne, Anthony Simeonov, Zechu Li, April Chan, Tao Chen, Abhishek Gupta, and Pulkit Agrawal.
\newblock Reconciling reality through simulation: A real-to-sim-to-real approach for robust manipulation.
\newblock \emph{arXiv}, 2024.

\bibitem[Traag et~al.()Traag, Waltman, and Van~Eck]{traag2019louvain}
Vincent~A Traag, Ludo Waltman, and Nees~Jan Van~Eck.
\newblock From louvain to leiden: guaranteeing well-connected communities.
\newblock \emph{Scientific reports}.

\bibitem[Vecchio et~al.(2024)Vecchio, Sortino, Palazzo, and Spampinato]{vecchio2024matfuse}
Giuseppe Vecchio, Renato Sortino, Simone Palazzo, and Concetto Spampinato.
\newblock Matfuse: controllable material generation with diffusion models.
\newblock In \emph{CVPR}, 2024.

\bibitem[Verbin et~al.(2022)Verbin, Hedman, Mildenhall, Zickler, Barron, and Srinivasan]{verbin2022ref}
Dor Verbin, Peter Hedman, Ben Mildenhall, Todd Zickler, Jonathan~T Barron, and Pratul~P Srinivasan.
\newblock Ref-nerf: Structured view-dependent appearance for neural radiance fields.
\newblock In \emph{2022 IEEE/CVF Conference on Computer Vision and Pattern Recognition (CVPR)}, pages 5481--5490. IEEE, 2022.

\bibitem[Wang et~al.(2021)Wang, Liu, Liu, Theobalt, Komura, and Wang]{wang2021neus}
Peng Wang, Lingjie Liu, Yuan Liu, Christian Theobalt, Taku Komura, and Wenping Wang.
\newblock Neus: Learning neural implicit surfaces by volume rendering for multi-view reconstruction.
\newblock \emph{arXiv}, 2021.

\bibitem[Wang et~al.(2024)Wang, Yuan, Zhang, Chen, Wang, Zhang, Shen, Zhao, and Zhou]{wang2024videocomposer}
Xiang Wang, Hangjie Yuan, Shiwei Zhang, Dayou Chen, Jiuniu Wang, Yingya Zhang, Yujun Shen, Deli Zhao, and Jingren Zhou.
\newblock Videocomposer: Compositional video synthesis with motion controllability.
\newblock \emph{NeurIPS}, 2024.

\bibitem[Wang et~al.(2023)Wang, Han, Habermann, Daniilidis, Theobalt, and Liu]{wang2023neus2}
Yiming Wang, Qin Han, Marc Habermann, Kostas Daniilidis, Christian Theobalt, and Lingjie Liu.
\newblock Neus2: Fast learning of neural implicit surfaces for multi-view reconstruction, 2023.

\bibitem[Wen et~al.(2024)Wen, Zhao, Ren, Schwing, and Wang]{wen2024gomavatar}
Jing Wen, Xiaoming Zhao, Zhongzheng Ren, Alexander~G. Schwing, and Shenlong Wang.
\newblock Gomavatar: Efficient animatable human modeling from monocular video using gaussians-on-mesh.
\newblock \emph{arXiv}, 2024.

\bibitem[Wu et~al.(2022)Wu, Wang, Pan, Xu, Theobalt, Liu, and Lin]{wu2022voxurf}
Tong Wu, Jiaqi Wang, Xingang Pan, Xudong Xu, Christian Theobalt, Ziwei Liu, and Dahua Lin.
\newblock Voxurf: Voxel-based efficient and accurate neural surface reconstruction.
\newblock \emph{arXiv preprint arXiv:2208.12697}, 2022.

\bibitem[Xia et~al.(2024)Xia, Lin, Ma, and Wang]{xia2024video2game}
Hongchi Xia, Zhi-Hao Lin, Wei-Chiu Ma, and Shenlong Wang.
\newblock Video2game: Real-time interactive realistic and browser-compatible environment from a single video.
\newblock In \emph{CVPR}, 2024.

\bibitem[Xie et~al.(2024)Xie, Zong, Qiu, Li, Feng, Yang, and Jiang]{xie2024physgaussianphysicsintegrated3dgaussians}
Tianyi Xie, Zeshun Zong, Yuxing Qiu, Xuan Li, Yutao Feng, Yin Yang, and Chenfanfu Jiang.
\newblock Physgaussian: Physics-integrated 3d gaussians for generative dynamics, 2024.

\bibitem[Xiong et~al.(2023)Xiong, Ma, and Urtasun]{ultralidar}
Yuwen Xiong, Jingkang Ma, Wei-Chiu~Wang, and Raquel Urtasun.
\newblock Ultralidar: Learning compact representations for lidar completion and generation.
\newblock \emph{CVPR}, 2023.

\bibitem[Yang et~al.(2023{\natexlab{a}})Yang, Zhang, Li, Zou, Li, and Gao]{yang2023set}
Jianwei Yang, Hao Zhang, Feng Li, Xueyan Zou, Chunyuan Li, and Jianfeng Gao.
\newblock Set-of-mark prompting unleashes extraordinary visual grounding in gpt-4v.
\newblock \emph{arXiv}, 2023{\natexlab{a}}.

\bibitem[Yang et~al.(2024)Yang, Ivanovic, Litany, Weng, Kim, Li, Che, Xu, Fidler, Pavone, and Wang]{yang2023emernerf}
Jiawei Yang, Boris Ivanovic, Or Litany, Xinshuo Weng, Seung~Wook Kim, Boyi Li, Tong Che, Danfei Xu, Sanja Fidler, Marco Pavone, and Yue Wang.
\newblock Emernerf: Emergent spatial-temporal scene decomposition via self-supervision.
\newblock In \emph{ICLR}, 2024.

\bibitem[Yang et~al.(2023{\natexlab{b}})Yang, Du, Ghasemipour, Tompson, Schuurmans, and Abbeel]{yang2023learning}
Mengjiao Yang, Yilun Du, Kamyar Ghasemipour, Jonathan Tompson, Dale Schuurmans, and Pieter Abbeel.
\newblock Learning interactive real-world simulators.
\newblock \emph{arXiv}, 2023{\natexlab{b}}.

\bibitem[Yang et~al.(2020{\natexlab{a}})Yang, Chai, Anguelov, Zhou, Sun, Erhan, Rafferty, and Kretzschmar]{yang2020surfelgan}
Zhenpei Yang, Yuning Chai, Dragomir Anguelov, Yin Zhou, Pei Sun, Dumitru Erhan, Sean Rafferty, and Henrik Kretzschmar.
\newblock Surfelgan: Synthesizing realistic sensor data for autonomous driving.
\newblock In \emph{CVPR}, 2020{\natexlab{a}}.

\bibitem[Yang et~al.(2020{\natexlab{b}})Yang, Chai, Anguelov, Zhou, Sun, Erhan, Rafferty, and Kretzschmar]{yang2020surfelgansynthesizingrealisticsensor}
Zhenpei Yang, Yuning Chai, Dragomir Anguelov, Yin Zhou, Pei Sun, Dumitru Erhan, Sean Rafferty, and Henrik Kretzschmar.
\newblock Surfelgan: Synthesizing realistic sensor data for autonomous driving, 2020{\natexlab{b}}.

\bibitem[Yang et~al.(2023{\natexlab{c}})Yang, Chen, Wang, Manivasagam, Ma, Yang, and Urtasun]{yang2023unisim}
Ze Yang, Yun Chen, Jingkang Wang, Sivabalan Manivasagam, Wei-Chiu Ma, Anqi~Joyce Yang, and Raquel Urtasun.
\newblock Unisim: A neural closed-loop sensor simulator.
\newblock \emph{CVPR}, 2023{\natexlab{c}}.

\bibitem[Yariv et~al.(2021)Yariv, Gu, Kasten, and Lipman]{yariv2021volume}
Lior Yariv, Jiatao Gu, Yoni Kasten, and Yaron Lipman.
\newblock Volume rendering of neural implicit surfaces.
\newblock \emph{NeurIPS}, 2021.

\bibitem[Yariv et~al.(2023)Yariv, Hedman, Reiser, Verbin, Srinivasan, Szeliski, Barron, and Mildenhall]{yariv2023bakedsdf}
Lior Yariv, Peter Hedman, Christian Reiser, Dor Verbin, Pratul~P Srinivasan, Richard Szeliski, Jonathan~T Barron, and Ben Mildenhall.
\newblock Bakedsdf: Meshing neural sdfs for real-time view synthesis.
\newblock In \emph{SIGGRAPH Conference}, 2023.

\bibitem[Yi et~al.(2018)Yi, Huang, Liu, Kalogerakis, Su, and Guibas]{yi2018deep}
Li Yi, Haibin Huang, Difan Liu, Evangelos Kalogerakis, Hao Su, and Leonidas Guibas.
\newblock Deep part induction from articulated object pairs.
\newblock \emph{arXiv}, 2018.

\bibitem[Yu et~al.(2022{\natexlab{a}})Yu, Chen, Antic, Peng, Bhattacharyya, Niemeyer, Tang, Sattler, and Geiger]{Yu2022SDFStudio}
Zehao Yu, Anpei Chen, Bozidar Antic, Songyou Peng, Apratim Bhattacharyya, Michael Niemeyer, Siyu Tang, Torsten Sattler, and Andreas Geiger.
\newblock Sdfstudio: A unified framework for surface reconstruction, 2022{\natexlab{a}}.

\bibitem[Yu et~al.(2022{\natexlab{b}})Yu, Peng, Niemeyer, Sattler, and Geiger]{yu2022monosdf}
Zehao Yu, Songyou Peng, Michael Niemeyer, Torsten Sattler, and Andreas Geiger.
\newblock Monosdf: Exploring monocular geometric cues for neural implicit surface reconstruction.
\newblock \emph{arXiv}, 2022{\natexlab{b}}.

\bibitem[Yu et~al.(2024)Yu, Sattler, and Geiger]{yu2024gaussian}
Zehao Yu, Torsten Sattler, and Andreas Geiger.
\newblock Gaussian opacity fields: Efficient and compact surface reconstruction in unbounded scenes.
\newblock \emph{arXiv}, 2024.

\bibitem[Yuan et~al.()Yuan, Murali, Mousavian, and Fox]{yuanm2t2}
Wentao Yuan, Adithyavairavan Murali, Arsalan Mousavian, and Dieter Fox.
\newblock M2t2: Multi-task masked transformer for object-centric pick and place.
\newblock In \emph{7th Annual Conference on Robot Learning}.

\bibitem[Ze et~al.(2024)Ze, Zhang, Zhang, Hu, Wang, and Xu]{ze20243d}
Yanjie Ze, Gu Zhang, Kangning Zhang, Chenyuan Hu, Muhan Wang, and Huazhe Xu.
\newblock 3d diffusion policy.
\newblock \emph{arXiv preprint arXiv:2403.03954}, 2024.

\bibitem[Zhai et~al.(2024)Zhai, Shen, Chen, Wang, Wang, Wang, Guan, and Wang]{zhai2024physical}
Albert~J Zhai, Yuan Shen, Emily~Y Chen, Gloria~X Wang, Xinlei Wang, Sheng Wang, Kaiyu Guan, and Shenlong Wang.
\newblock Physical property understanding from language-embedded feature fields.
\newblock In \emph{CVPR}, 2024.

\bibitem[Zhang et~al.(2021)Zhang, Fei, Song, and Feng]{zhang2021improved}
Jicun Zhang, Jiyou Fei, Xueping Song, and Jiawei Feng.
\newblock An improved louvain algorithm for community detection.
\newblock \emph{Mathematical Problems in Engineering}, 2021.

\bibitem[Zhang et~al.(2018)Zhang, Isola, Efros, Shechtman, and Wang]{zhang2018perceptual}
Richard Zhang, Phillip Isola, Alexei~A Efros, Eli Shechtman, and Oliver Wang.
\newblock The unreasonable effectiveness of deep features as a perceptual metric.
\newblock In \emph{CVPR}, 2018.

\bibitem[Zhang et~al.(2024)Zhang, Yu, Wu, Feng, Zheng, Snavely, Wu, and Freeman]{zhang2024physdreamer}
Tianyuan Zhang, Hong-Xing Yu, Rundi Wu, Brandon~Y. Feng, Changxi Zheng, Noah Snavely, Jiajun Wu, and William~T. Freeman.
\newblock {PhysDreamer}: Physics-based interaction with 3d objects via video generation.
\newblock In \emph{ECCV}, 2024.

\bibitem[Zhang et~al.(2023)Zhang, Wei, Jiang, Zhang, Zuo, and Tian]{zhang2023controlvideo}
Yabo Zhang, Yuxiang Wei, Dongsheng Jiang, Xiaopeng Zhang, Wangmeng Zuo, and Qi Tian.
\newblock Controlvideo: Training-free controllable text-to-video generation.
\newblock \emph{arXiv}, 2023.

\bibitem[Zhao et~al.(2021)Zhao, Han, Zhang, Xu, and Cheng]{zhao2021deep}
Kai Zhao, Qi Han, Chang-Bin Zhang, Jun Xu, and Ming-Ming Cheng.
\newblock Deep hough transform for semantic line detection.
\newblock \emph{IEEE Transactions on Pattern Analysis and Machine Intelligence}, 44\penalty0 (9):\penalty0 4793--4806, 2021.

\bibitem[Zhao et~al.(2024)Zhao, Wu, Huang, Zhi, Zhao, Wang, and Gao]{zhao2024surfel}
Yiqun Zhao, Chenming Wu, Binbin Huang, Yihao Zhi, Chen Zhao, Jingdong Wang, and Shenghua Gao.
\newblock Surfel-based gaussian inverse rendering for fast and relightable dynamic human reconstruction from monocular video.
\newblock \emph{arXiv preprint arXiv:2407.15212}, 2024.

\bibitem[Zheng et~al.(2024)Zheng, Peng, Yang, Shen, Li, Liu, Zhou, Li, and You]{opensora}
Zangwei Zheng, Xiangyu Peng, Tianji Yang, Chenhui Shen, Shenggui Li, Hongxin Liu, Yukun Zhou, Tianyi Li, and Yang You.
\newblock Open-sora: Democratizing efficient video production for all, 2024.

\bibitem[Zyrianov et~al.(2022)Zyrianov, Zhu, and Wang]{zyrianov2022learning}
Vlas Zyrianov, Xiyue Zhu, and Shenlong Wang.
\newblock Learning to generate realistic lidar point clouds.
\newblock In \emph{ECCV}, 2022.

\bibitem[Zyrianov et~al.(2024)Zyrianov, Che, Liu, and Wang]{zyrianov2024lidardmgenerativelidarsimulation}
Vlas Zyrianov, Henry Che, Zhijian Liu, and Shenlong Wang.
\newblock Lidardm: Generative lidar simulation in a generated world, 2024.

\end{thebibliography}
